\pdfoutput=1

\documentclass[11pt]{article}

\usepackage[preprint]{acl}

\usepackage{times}
\usepackage{latexsym}

\usepackage[T1]{fontenc}

\usepackage[utf8]{inputenc}

\usepackage{microtype}

\usepackage{inconsolata}

\usepackage{graphicx}

\usepackage{subfig}
\usepackage{booktabs} 
\usepackage{multirow}
\usepackage{colortbl}
\usepackage{xspace}
\usepackage{tabularx}
\usepackage{longtable}
\usepackage{amsmath}

\usepackage[capitalize]{cleveref}
\crefname{section}{Sec.}{Secs.}
\Crefname{section}{Section}{Sections}
\Crefname{table}{Table}{Tables}
\crefname{table}{Tab.}{Tabs.}

\usepackage{enumitem}
\setenumerate{itemsep=0pt,partopsep=0pt,parsep=0pt,topsep=2pt}

\newcommand{\name}{QLLM\xspace}
\newcommand{\llamagrad}{LLaMA3-8B-inst-1048K \xspace}
\newcommand{\llama}{QuickLLaMA\xspace}

\newcommand{\lookup}{Query-aware Context Lookup\xspace}
\newcommand{\fullname}{Query-aware Inference for LLMs\xspace}
\newcommand{\infllm}{ILM\xspace}
\newcommand{\needle}{Needle-in-a-Haystack\xspace}
\newcommand{\basellama}{LLaMA3-8B-inst\xspace}
\newcommand{\basemistral}{Mistral-7B-inst-v0.2\xspace}
\title{\raisebox{-2mm}{\includegraphics[width=0.9cm, trim=0 50 0 50, clip]{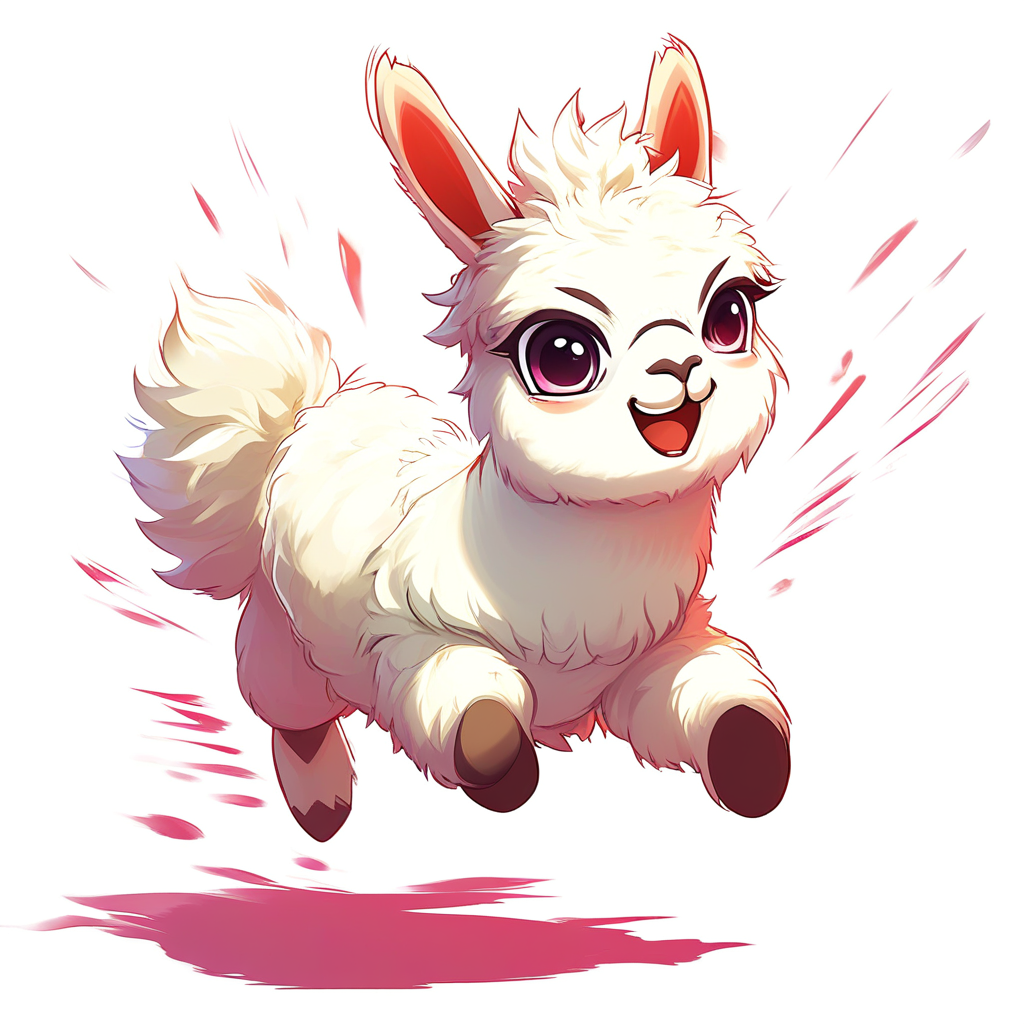}}\llama: Query-aware Inference Acceleration \\ for Large Language Models}

\author{
 \textbf{Jingyao Li\textsuperscript{1}},
 \textbf{Han Shi\textsuperscript{2}},
 \textbf{Xin Jiang\textsuperscript{2}},
 \textbf{Zhenguo Li\textsuperscript{2}},
 \textbf{Hong Xu\textsuperscript{1}},
 \textbf{Jiaya Jia\textsuperscript{1,3}}
\\
\\
  \textsuperscript{1}CUHK,
  \textsuperscript{2}Huawei Noah’s Ark Lab,
  \textsuperscript{3}SmartMore
}

\begin{document}

\maketitle

\begin{abstract}
The capacity of Large Language Models (LLMs) to comprehend and reason over long contexts is pivotal for advancements in diverse fields. Yet, they still struggle with identifying relevant contexts and memory searching. To address this issue, we introduce \fullname (\name), a system designed to process extensive sequences akin to human cognition. By focusing on memory data relevant to a given query, \name accurately captures pertinent information within a fixed window size and provides precise answers to queries. It requires no additional training and can be seamlessly integrated with any LLMs. Using LLaMA3 (QuickLLaMA), \name can read \emph{Harry Potter} within 30 seconds and accurately answer related questions. On widely recognized benchmarks, \name improved performance by 7.17\% compared to the current state-of-the-art on LLaMA3 and by 3.26\% on Mistral on the $\infty$-bench. In the \needle and BABILong task, \name improved upon the current SOTA by 7.0\% and 6.1\%. Our code can be found in \href{https://github.com/dvlab-research/Q-LLM}{https://github.com/dvlab-research/Q-LLM}


\end{abstract}

\section{Introduction}
The ability to understand and reason over broad contexts has always been a long-term research focus of Large Language Models (LLMs)~\cite{long-survey}. LLM-driven agents need to process ongoing information from external sources, which requires a strong ability to manage lengthy sequences~\cite{robocoder,zheng2024cape}; An ideal ChatBot assistant should be able to operate consistently over the content of conversations spanning recent days~\cite{gpt4}. Other tasks such as summarizing and answering questions based on books, reports, and documents, as well as generating code at the repository level, also require the capability to handle long context sequences~\cite{longbench,infinitebench}.

\begin{figure*}[p]
  \centering
    \includegraphics[width=0.99\linewidth, trim=0 210 60 0, clip]{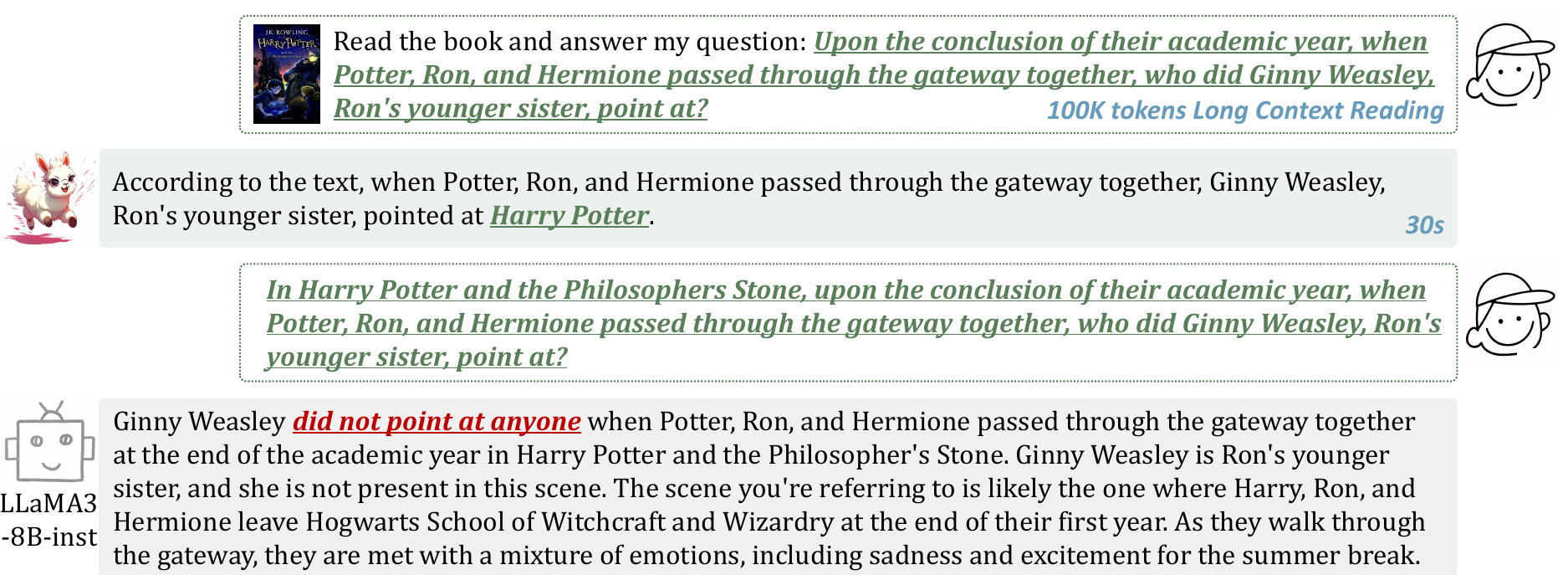}
    \\
    \includegraphics[width=0.99\linewidth, trim=0 210 60 0, clip]{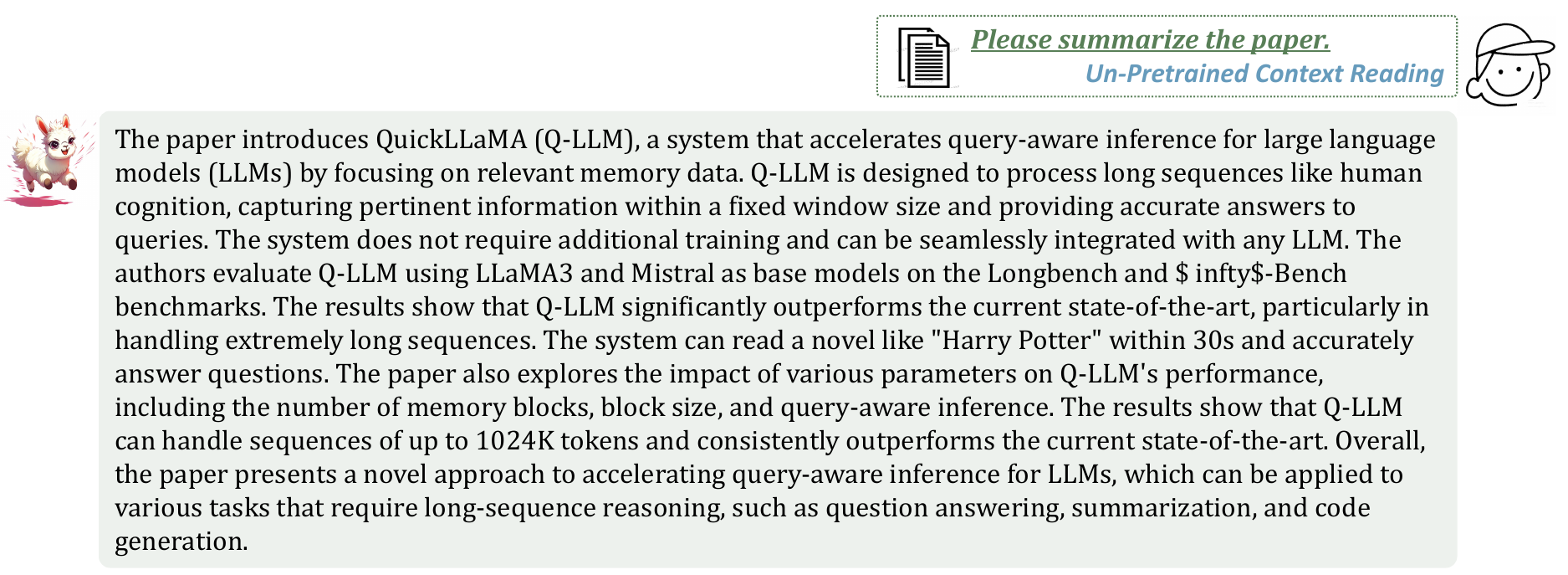}
    \\
    \includegraphics[width=0.99\linewidth, trim=0 190 60 0, clip]{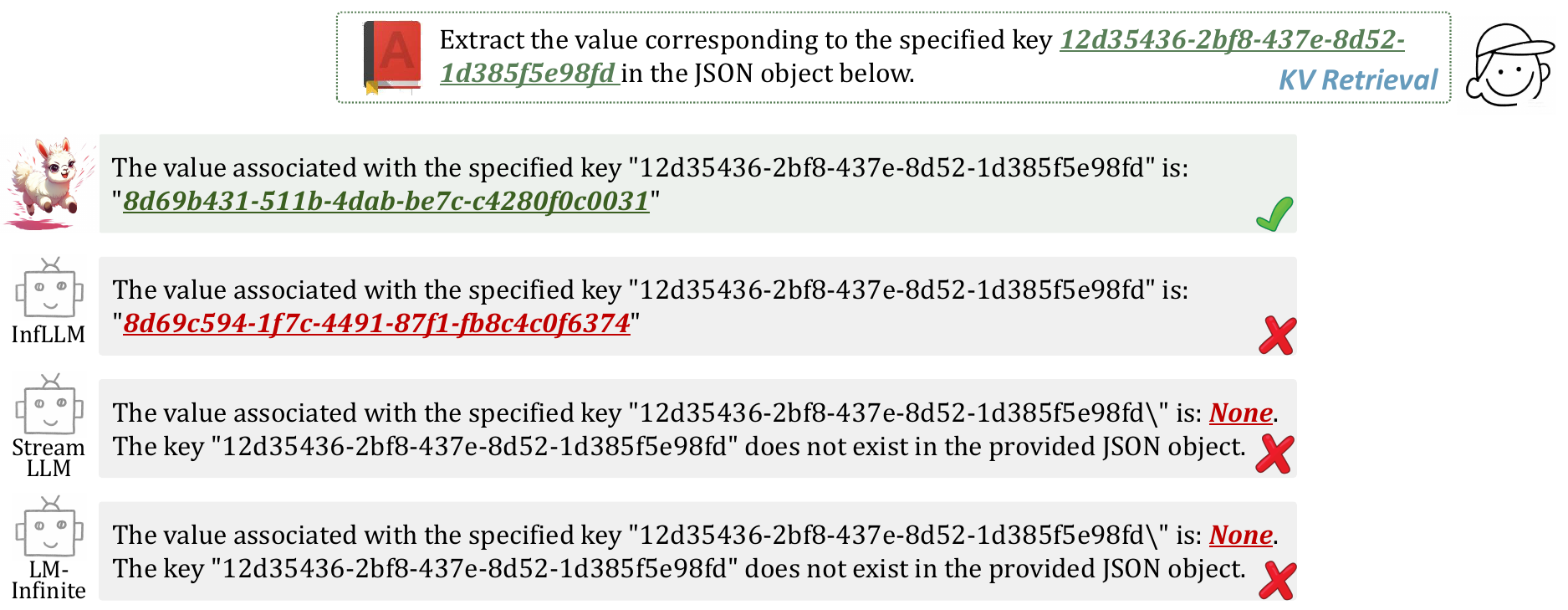}
    \\
    \includegraphics[width=0.99\linewidth, trim=0 260 60 0, clip]{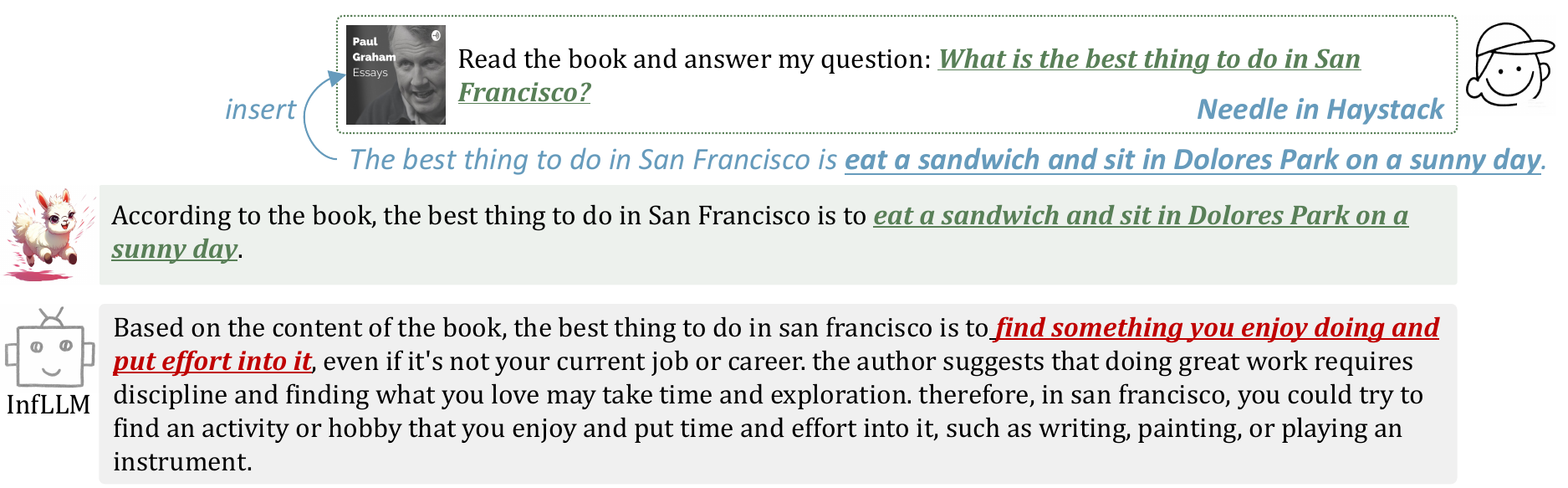}
    \caption{Examples of our \llama-8B (1) reading long context containing 100K tokens, (2) reading our paper that has not be seen in the pretrained dataset, (3) retrieving value in long key-value pairs and (4) retrieving in \needle task. More examples and comparisons with the SOTAs are provided in \cref{sec:a-examples}.} 
    \label{fig:exp}
\end{figure*}

\begin{figure*}[t]
  \centering
        \includegraphics[width=0.99\linewidth, trim=0 10 0 10, clip]{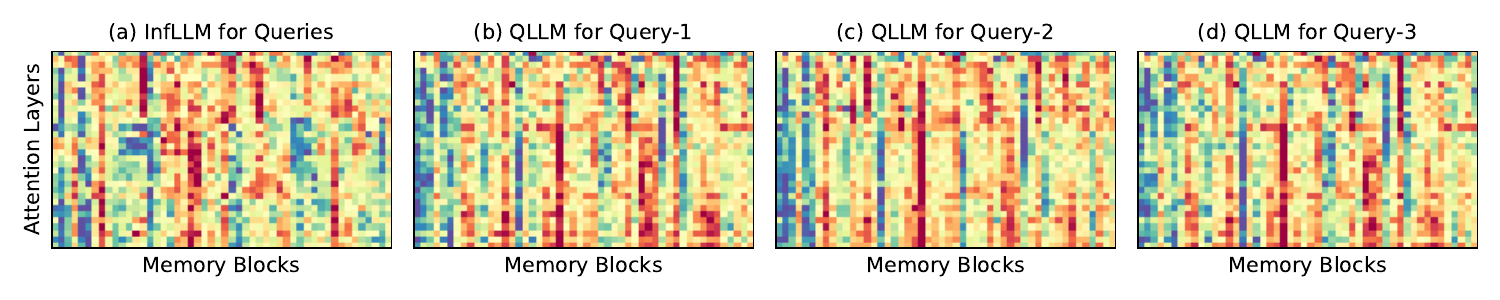}
    \caption{This is an example from the \(\infty\)-Bench. Three questions were posed about the same long book: (1) \emph{Which among Annalisa, Seb, Peyton, and Gannonmarie is not Mrs. Bronwyn's child?} (2) \emph{What's the name of the Bronwyns' summer home?} (3) \emph{Who among Mrs. Bronwyn, Mrs. Deandra, Rosemarie, and Cael is the final to perish?} We present the score heatmap of the first 50 memory blocks. The methods used include (a) the consistent results from InfLLM for all three queries, and (b-d) the query-aware results from \name.}
    \label{fig:hotmap}
\end{figure*}

Yet, due to the challenges posed by unobserved extensive inputs~\cite{infinite-llm} and distracting, noisy contexts~\cite{DBLP:journals/corr/abs-2307-03172,DBLP:journals/corr/abs-2307-03170}, most LLMs that are pre-trained on sequences comprising a few thousand tokens struggle to generalize on longer sequences, resulting in unsatisfactory performance~\cite{alibi,DBLP:journals/corr/abs-2312-17044}. Some contemporary studies make use of sliding windows to disregard distant contexts, thereby ensuring that the length of the sequence do not surpass the LLMs' maximum capacity~\cite{stream-llm,infinite-llm} and incorporate block-level context memory, which opts pertinent information from memory to disregard irrelevant disturbances~\cite{infllm}. However, the memory to be focused on should differ according to the specific query requirements. Yet, for distinct queries, InfLLM~\cite{infllm} exhibits identical focal points when reading the long context, as shown in~\cref{fig:hotmap}.


To address these challenges, we design \fullname (\name), which processes extensive sequences in a manner similar to human cognition. Humans, when interpreting text, initially examine the question, and then seek the answer within the context, keeping the query in mind. This idea forms the foundation of our \lookup strategy. Only memory data pertinent to the query is chosen for each computational step, disregarding unrelated distractions. As a result, LLMs can capture pertinent information within a fixed window size and provide precise answers to queries. \name doesn't require extra training and can be seamlessly integrated with any LLMs. 

We assess the performance of \name by utilizing \basellama~\cite{llama3} (\llama) and \basemistral~\cite{mistral} as foundational models. These base models are pre-trained on sequences that do not exceed $8$K tokens. Instead, our \llama can read \emph{Harry Potter} with 100K tokens within half a minute on a single A800 GPU and accurately answer the questions, as shown in \cref{fig:exp}. We employ several widely recognized benchmarks, namely Longbench~\cite{longbench}, $\infty$-Bench~\cite{infinitebench}, and \needle Benchmark. Specifically, with a context window of 512, \name improved by 7.17\% compared to the current SOTA on LLaMA3, and by 3.26\% on Mistral on the $\infty$-bench. In the \needle task, \name improved upon the current SOTA by by 7.0\% on Mistral and achieves 100\% on LLaMA3. In the BABILong task, \name improved upon the current SOTA by 6.1\%. We have extended the input sequence to contexts of 1048K length, further demonstrating our model's capability in handling extremely long sequences.

\begin{figure*}[t]
  \centering
        \includegraphics[width=0.99\linewidth, trim=160 160 125 170, clip]{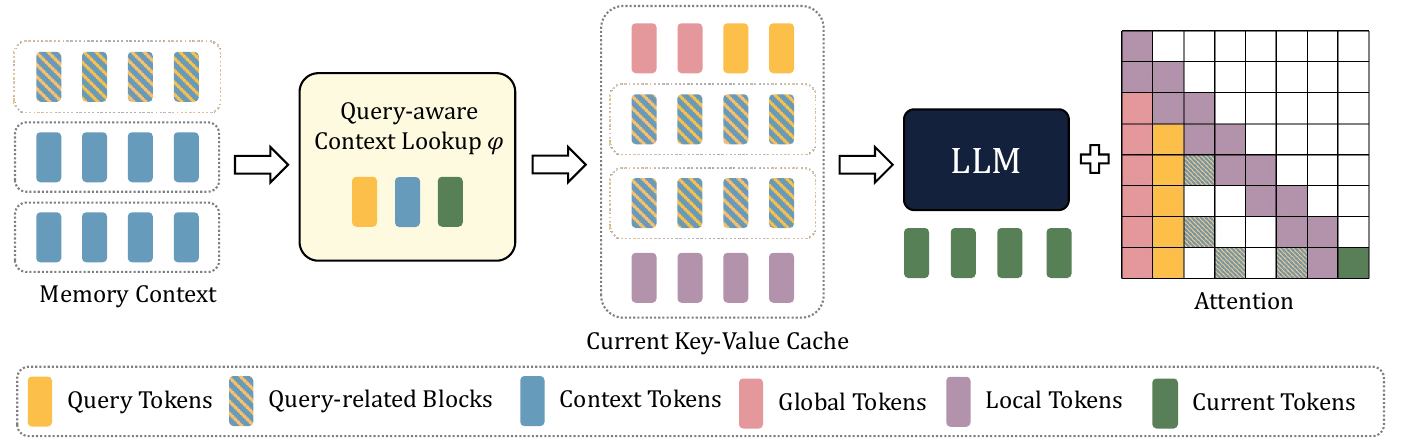}
    \caption{The illustration of our \name framework. The input from the memory context is partitioned into memory blocks, which are searched by \lookup for query-related blocks. The current key-value cache comprises global tokens, query tokens, query-related blocks, and local tokens. Together, these form a new context window that, along with current tokens, is fed into the LLM.}
    \label{fig:framework}
\end{figure*}

\section{Related Works}
\label{sec:a-related-works}
\noindent\textbf{Efficient Context Computation.}
The computational and memory demands of LLM training often limit it to short sequences. Using LLMs directly on long sequences presents challenges such as out-of-domain issues and distractions from lengthy and noisy inputs~\cite{infinite-llm, DBLP:journals/corr/abs-2307-03170, snapkv}. As a result, context length extrapolation has emerged as a method to extend LLMs' sequence length without additional training. 
Early approaches have designed new relative positional encoding mechanisms during pre-training~\cite{alibi,xpos}.
The following research has focused on the extensively adopted rotary position embedding (RoPE)~\cite{rope}, suggesting extending the length by interpolating positions to introduce non-integer positions~\cite{pi,yarn,DBLP:journals/corr/abs-2401-01325,DBLP:journals/corr/abs-2310-16450}. 
To process extremely long sequences, Stream-LLM~\citep{stream-llm} and LM-Infinite\citep{infinite-llm} utilize the sliding window attention mechanism and discard distant contexts. Additionally, InfLLM~\cite{infllm} leverages a context memory to furnish LLMs with pertinent contextual information. Yet, the objective of these models during long-text reading is inherently ambiguous, and it can become distracting when reading extensive articles. In this work, we introduce the \lookup mechanism, enabling the model to effectively retrieve information relevant to the query from lengthy texts.

\noindent\textbf{Context Length Extrapolation.}
The computational complexity of attention layers, which grows quadratically, is a significant bottleneck restricting LLMs' capability to handle lengthy sequences. Consequently, numerous researchers have devised efficient attention mechanisms, including sparse attention~\cite{bigbird,longformer,DBLP:journals/corr/abs-1904-10509,etc,DBLP:journals/corr/abs-1912-11637}, approximate attention computations using kernel functions~\cite{reformer,linformer,linear-transformer}, and replacing attention layers with state-space models of linear complexity~\cite{s4,Mamba}. These approaches necessitate architectural modifications, requiring retraining of the models. Concurrently, many scholars have tackled this challenge from an infrastructural angle by optimizing the memory usage of attention computations to mitigate the computational resource requirements of the model~\cite{flashattention,flashattention-2,flashdecoding++,DBLP:journals/corr/abs-1911-02150,vllm}. Given the training-free nature of our method, it can be seamlessly integrated to further expedite LLM inference.

\noindent\textbf{Memory-based Approaches.}
Memory networks have been extensively researched for decades and have demonstrated effectiveness in enhancing models with additional information storage capabilities~\cite{NTM,MemNet,DBLP:conf/nips/SukhbaatarSWF15,DBLP:conf/emnlp/MillerFDKBW16}. With the rise of pre-trained models, memory layers have gradually found application in the training stage of recurrent transformer layers, enabling models to recursively process long sequences~\cite{transformer-xl,DBLP:conf/iclr/RaePJHL20,DBLP:conf/iclr/KhandelwalLJZL20,memorizing-transformer,unlimiformer}. These approaches segment sequences, encoding each segments individually, and utilize memory to retain context information from preceding segments. Yet, they necessitate architectural modifications and are typically incorporated during the pre-training phase. In contrast, our objective is to explore the intrinsic properties of LLMs and introduce a training-free \lookup mechanism for long-text comprehension.

\section{Methods}
In this section, we introduce the overall framework of \fullname (\name) in \cref{sec:framework}, as depicted in \cref{fig:framework}. Then, we propose the preliminary memory block in \cref{sec:block} and our proposed \lookup in \cref{sec:lookup}.

\subsection{Overall Framework}
\label{sec:framework}
\begin{figure*}[t]
  \centering
        \includegraphics[width=0.99\linewidth, trim=125 125 110 260, clip]{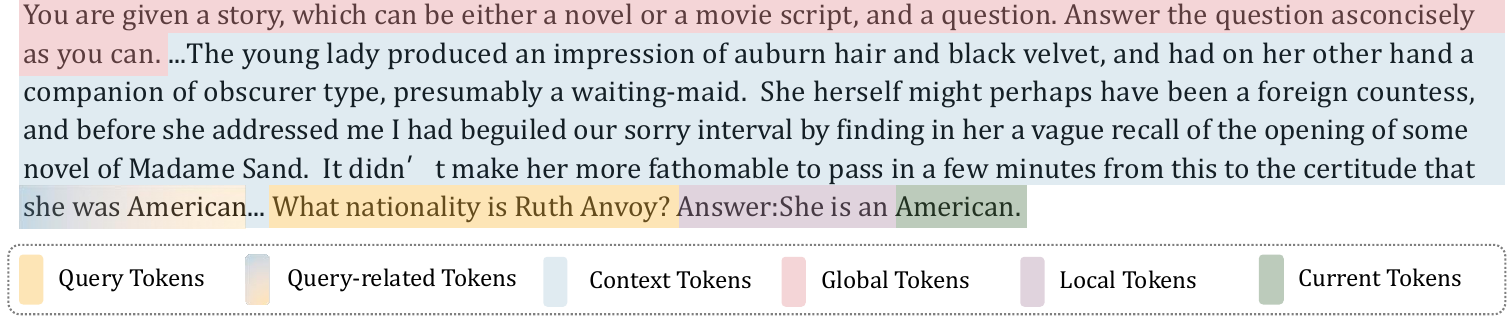}
    \caption{An example from Long-Bench. Global tokens include system prompts and task description. Query tokens represent the query of the user. Context tokens indicate the context stored in the context memory. We search query-related tokens from them, local tokens are the nearest tokens to the current token.}
    \label{fig:example}
\end{figure*}

The primary challenges in expanding the context window of LLMs arise from issues related to out-of-domain and distractions, which are a result of the extensive and noisy context. To tackle these challenges, we follow prior studies, which implement the sliding window attention mechanism~\cite{stream-llm,infinite-llm} and the context memory module ~\cite{infllm}. Additionally, we propose the \lookup strategy to find the query-related tokens from the context token. The past key-value vectors $\mathbf{P} = \{(\mathbf{k}_i, \mathbf{v}_i)\}_{i=1}^{l_P}$ consist of four composers: 
\begin{enumerate}
    \item Global tokens $\mathbf{G}$, including system prompts and task description, etc. 
    \item Query tokens $\mathbf{Q}$, the query of the user.
    \item Context tokens $\mathbf{C}$, the context stored in the context memory, consisting of multiple memory blocks.
    \item Local tokens $\mathbf{L}$, the nearest tokens to the current token.
\end{enumerate}
An example of these tokens in the prompt is shown in \cref{fig:example}. Given that all memory blocks are necessary to be maintained and most of them and seldom used, we adopt an offloading strategy, which stores most memory blocks in CPU memory. More details are in \cref{sec:cache}. We propose the \lookup strategy to find the query-related tokens $\mathbf{R}$ from the context tokens $\mathbf{C}$:
\begin{equation}
    \mathbf{R} = \phi(\mathbf{H}, \mathbf{C}, \mathbf{Q}),
\end{equation}
where $\phi(\cdot)$ refers to the lookup operation of context memory. We will detail the strategy in \cref{sec:lookup}. For each step, \name combines the global tokens, query tokens, query-related tokens, and local tokens to compose the current key-value cache.
\begin{equation}
\mathbf{M} = \text{Concat}(\mathbf{G}, \mathbf{Q}, \mathbf{R}, \mathbf{L}),
\end{equation}
Finally, the input parameters of the attention are:
\begin{equation}
\begin{aligned}
    \mathbf{A_q} &= \mathbf{P_q}\mathbf{H}, \\
    \mathbf{A_k} &=  \text{Concat}(\mathbf{M_k}, \mathbf{P_k}\mathbf{H}), \\
    \mathbf{A_v} &=  \text{Concat}(\mathbf{M_v}, \mathbf{P_v}\mathbf{H}),
\end{aligned}
\end{equation}
where $\mathbf{P_q}$, $\mathbf{P_k}$, and $\mathbf{P_v}$ are parameters in attention layers, $\mathbf{M_k}$ and $\mathbf{M_v}$ refer to the key and value vectors in the current key-value cache $\mathbf{M}$.

\begin{table*}[t]
  \centering
\small

\subfloat[$\infty$-Bench (214K tokens)]{
\setlength{\tabcolsep}{1mm}
\begin{tabular}{l|rrr >{\columncolor{teal!15}}r|rrr >{\columncolor{teal!15}}r|rrr >{\columncolor{teal!15}}r}
\toprule
Method & Infinite & Stream & \infllm & \cellcolor{white}\name & Infinite & Stream & \infllm & \cellcolor{white}\name & Infinite & Stream & \infllm & \cellcolor{white}\name \\
Context Window & \multicolumn{4}{c|}{512} & \multicolumn{4}{c|}{1K} & \multicolumn{4}{c}{2K} \\
\midrule
En.MC & 26.64 & 27.95 & 26.63 & \textbf{29.69} & 31.00 & 30.13 & 33.19 & \textbf{33.19} & 30.13 & 30.57 & 33.62 & \textbf{34.50} \\
Retrieve.PassKey & 3.40 & 3.40 & 100.0 & \textbf{100.0} & 3.40 & 3.40 & 100.0 & \textbf{100.0} & 3.40 & 3.40 & 100.0 & \textbf{100.0} \\
Retrieve.Number & 3.39 & 3.39 & 99.73 & \textbf{100.0} & 3.39 & 3.39 & 99.83 & \textbf{99.83} & 3.39 & 3.39 & \textbf{44.58} & 40.00 \\
Code.Debug & 31.22 & \textbf{32.74} & 31.02 & 31.22 & 35.79 & 37.56 & 38.58 & \textbf{38.58} & \textbf{35.79} & 32.74 & 35.03 & 34.52 \\
Math.Find & 17.71 & 17.14 & 24.77 & \textbf{25.93} & 16.57 & 17.43 & 27.71 & \textbf{28.86} & 15.43 & 16.29 & 28.29 & \textbf{28.29} \\
Retrieve.KV & 0.20 & 0.20 & 13.18 & \textbf{32.40} & 0.40 & 0.40 & 32.60 & \textbf{47.80} & 1.00 & 1.00 & 62.94 & \textbf{73.00} \\
\midrule
Average & 13.76 & 14.14 & 49.22 & \textbf{53.21} & 15.09 & 15.38 & 55.32 & \textbf{58.04} & 14.86 & 14.56 & 50.74 & \textbf{51.72} \\ 
\bottomrule
\end{tabular}
\label{tab:mistral-infinite}
}

\subfloat[Long-Bench (31K tokens)]{
\setlength{\tabcolsep}{1mm}
\begin{tabular}{l|rrr >{\columncolor{teal!15}}r|rrr >{\columncolor{teal!15}}r|rrr >{\columncolor{teal!15}}r}
\toprule
Method & Infinite & Stream & \infllm & \cellcolor{white}\name & Infinite & Stream & \infllm & \cellcolor{white}\name & Infinite & Stream & \infllm & \cellcolor{white}\name \\
Context Window & \multicolumn{4}{c|}{512} & \multicolumn{4}{c|}{1K} & \multicolumn{4}{c}{2K} \\
\midrule
NarrativeQA & 8.80 & 9.77 & 11.80 & \textbf{12.04} & 9.44 & 10.19 & 15.61 & \textbf{15.95} & 12.44 & 13.37 & 18.75 & \textbf{20.14} \\
Qasper & 9.19 & 9.45 & \textbf{16.13} & 15.45 & 10.93 & 10.73 & 19.15 & \textbf{19.24} & 14.58 & 15.04 & \textbf{20.78} & 19.97 \\
MultiFieldQA & 25.38 & 26.03 & 38.43 & \textbf{41.35} & 27.82 & 27.76 & 42.65 & \textbf{43.71} & 32.29 & 32.02 & 43.74 & \textbf{44.83} \\
HotpotQA & 19.68 & 20.46 & \textbf{28.19} & 27.32 & 22.16 & 21.91 & 32.47 & \textbf{34.47} & 23.21 & 23.70 & 34.66 & \textbf{36.53} \\
2WikiMQA & 12.27 & 12.63 & 13.70 & \textbf{15.22} & 13.85 & 13.32 & 16.14 & \textbf{16.57} & 17.13 & 17.51 & 17.99 & \textbf{19.97} \\
Musique & 6.45 & 6.55 & 11.38 & \textbf{12.99} & 7.91 & 7.64 & 14.74 & \textbf{15.27} & 9.81 & 11.30 & 12.16 & \textbf{17.05} \\
GovReport & 22.50 & 22.40 & \textbf{29.64} & 28.46 & 24.79 & 24.90 & \textbf{30.18} & 29.82 & 27.07 & 27.12 & \textbf{30.26} & 29.75 \\
QMSum & 18.74 & 18.93 & 21.55 & \textbf{21.69} & 19.23 & 19.19 & 22.03 & \textbf{22.27} & 19.67 & 19.52 & 21.55 & \textbf{22.36} \\
MultiNews & 23.23 & 23.28 & \textbf{25.19} & 24.95 & 25.51 & 25.41 & 26.15 & \textbf{26.39} & 25.95 & 26.10 & 26.71 & \textbf{26.84} \\
TREC & 38.00 & 39.50 & 45.50 & \textbf{47.50} & 30.50 & 29.00 & 48.00 & \textbf{49.50} & 31.00 & 28.25 & 47.50 & \textbf{48.25} \\
TriviaQA & 79.68 & 80.54 & 82.02 & \textbf{82.20} & \textbf{85.06} & 84.27 & 83.20 & 84.56 & \textbf{88.06} & 87.08 & 82.81 & 84.49 \\
SAMSum & 35.30 & 34.58 & 36.65 & \textbf{37.18} & 36.05 & 35.09 & \textbf{38.20} & 38.12 & 36.30 & 36.21 & 37.91 & \textbf{38.25} \\
PassageRetrieval & 4.40 & 5.54 & 13.29 & \textbf{25.04} & 7.92 & 7.92 & 25.67 & \textbf{31.04} & 18.21 & 18.46 & 40.29 & \textbf{49.67} \\
LCC & 50.06 & \textbf{51.59} & 50.14 & 48.61 & 50.94 & \textbf{53.27} & 50.83 & 51.10 & 52.20 & \textbf{54.95} & 54.59 & 54.52 \\
RepoBench-P & 47.38 & \textbf{48.04} & 42.92 & 41.32 & 48.85 & \textbf{51.31} & 41.75 & 43.21 & 47.36 & \textbf{47.60} & 45.08 & 45.90 \\
\midrule
Average & 25.18 & 25.59 & 29.30 & \textbf{30.22} & 26.40 & 26.44 & 31.80 & \textbf{32.70} & 28.56 & 28.77 & 33.55 & \textbf{35.06} \\
\bottomrule
\end{tabular}
\label{tab:mistral-long}
}
\caption{
    \label{tab:mistral}
    The results comparison based on Mistral-7B-inst-v0.2~\cite{mistral}. Our results are highlighted in teal and best results are indicated in bold. 
}
\end{table*}
\begin{table*}[t]
  \centering
\small

\subfloat[$\infty$-Bench (214K tokens)]{
\setlength{\tabcolsep}{0.6mm}
\begin{tabular}{l|r|rrr >{\columncolor{teal!15}}r|rrr >{\columncolor{teal!15}}r|rrr >{\columncolor{teal!15}}r}
\toprule
Method & LLaMA & Infinite & Stream & \infllm & \cellcolor{white}\name & Infinite & Stream & \infllm & \cellcolor{white}\name & Infinite & Stream & \infllm & \cellcolor{white}\name \\
Context Window & \multicolumn{1}{c|}{-1048K} & \multicolumn{4}{c|}{512} & \multicolumn{4}{c|}{1K} & \multicolumn{4}{c}{2K} \\
\midrule
En.MC & 31.0 & 37.12 & 34.93 & 37.77 & \textbf{40.17} & 37.12 & 34.93 & 37.99 & \textbf{40.17} & 36.24 & \textbf{37.12} & 33.19 & 34.50 \\
Retrieve.PassKey & 6.78 & 3.40 & 3.40 & 100.0 & \textbf{100.0} & 3.40 & 3.40 & 100.0 & \textbf{100.0} & 3.40 & 3.40 & 100.0 & \textbf{100.0} \\
Retrieve.Number & 6.78 & 3.39 & 3.39 & 96.61 & \textbf{98.98} & 3.39 & 3.39 & 40.68 & \textbf{41.19} & 3.39 & 3.39 & \textbf{28.64} & 27.63 \\
Code.Debug & 22.59 & 22.59 & 22.59 & 22.59 & \textbf{22.59} & 22.59 & 22.59 & 23.10 & \textbf{23.86} & \textbf{24.11} & 22.84 & 22.59 & 23.10 \\
Math.Find & 34.29 & 20.86 & 19.71 & 29.23 & \textbf{30.70} & 20.86 & 19.71 & \textbf{32.29} & 31.14 & 18.00 & 16.86 & 26.86 & \textbf{27.37} \\
Retrieve.KV & 6.2 & 0.80 & 0.80 & 24.40 & \textbf{61.20} & 0.80 & 0.80 & 57.20 & \textbf{70.80} & 1.80 & 1.80 & 80.80 & \textbf{84.00} \\
\midrule
Average & 17.94 & 14.69 & 14.14 & 51.77 & \textbf{58.94} & 14.69 & 14.14 & 48.54 & \textbf{51.19} & 14.49 & 14.23 & 48.68 & \textbf{49.43} \\
\bottomrule
\end{tabular}
\label{tab:llama3-infinite}
}

\subfloat[Long-Bench (31K tokens)]{
\setlength{\tabcolsep}{0.6mm}
\begin{tabular}{l|r|rrr >{\columncolor{teal!15}}r|rrr >{\columncolor{teal!15}}r|rrr >{\columncolor{teal!15}}r}
\toprule
Method & LLaMA & Infinite & Stream & \infllm & \cellcolor{white}\name & Infinite & Stream & \infllm & \cellcolor{white}\name & Infinite & Stream & \infllm & \cellcolor{white}\name \\
Context Window & \multicolumn{1}{c|}{-1048K} & \multicolumn{4}{c|}{512} & \multicolumn{4}{c|}{1K} & \multicolumn{4}{c}{2K} \\
\midrule
NarrativeQA & 23.78 & 14.50 & 14.56 & 19.28 & \textbf{19.29} & 14.50 & 14.56 & 19.90 & \textbf{20.50} & 16.47 & 15.12 & 19.41 & \textbf{25.60} \\
Qasper & 21.22 & 21.06 & 20.72 & 26.08 & \textbf{26.58} & 21.06 & 20.72 & \textbf{32.35} & 31.47 & 32.01 & 31.72 & \textbf{41.27} & 39.12 \\
MultiFieldQA & 39.89 & 25.66 & 25.79 & 36.01 & \textbf{40.12} & 25.66 & 25.79 & 41.46 & \textbf{46.44} & 31.63 & 30.99 & 45.89 & \textbf{48.30} \\
HotpotQA & 17.16 & 31.95 & 32.84 & 41.42 & \textbf{42.34} & 31.95 & 32.84 & 43.75 & \textbf{49.15} & 34.73 & 35.26 & 44.97 & \textbf{49.91} \\
2WikiMQA & 18.11 & 24.72 & 24.28 & 28.44 & \textbf{29.63} & 24.72 & 24.28 & 30.83 & \textbf{31.53} & 29.22 & 30.59 & 36.27 & \textbf{39.63} \\
Musique & 10.39 & 12.72 & 13.62 & 17.48 & \textbf{18.75} & 12.72 & 13.62 & 21.26 & \textbf{23.95} & 13.50 & 13.64 & 19.73 & \textbf{25.03} \\
GovReport & 33.76 & 26.25 & 25.93 & \textbf{29.26} & 26.83 & 26.25 & 25.93 & \textbf{30.44} & 28.73 & 27.84 & 27.83 & \textbf{30.68} & 29.80 \\
QMSum & 23.38 & 19.38 & \textbf{19.42} & 19.10 & 19.02 & 19.38 & 19.42 & 20.30 & \textbf{20.62} & 19.91 & 20.14 & 21.36 & \textbf{22.23} \\
MultiNews & 27.68 & 26.42 & 26.57 & \textbf{26.63} & 25.22 & 26.42 & 26.57 & \textbf{27.46} & 26.85 & 27.36 & 27.37 & \textbf{27.87} & 27.85 \\
TriviaQA & 87.76 & 82.46 & \textbf{82.47} & 80.81 & 77.65 & 82.46 & 82.47 & \textbf{85.11} & 85.04 & \textbf{88.07} & 87.35 & 88.03 & 87.70 \\
SAMSum & 41.89 & 38.28 & 37.91 & 38.73 & \textbf{38.83} & 38.28 & 37.91 & 39.59 & \textbf{40.40} & \textbf{36.93} & 35.97 & 34.86 & 34.97 \\ 
PassageRetrieval & 51.83 & 13.75 & 13.25 & 19.50 & \textbf{35.00} & 13.75 & 13.25 & 58.75 & \textbf{69.00} & 23.50 & 23.50 & 85.25 & \textbf{88.00} \\
LCC & 43.79 & \textbf{54.17} & 53.88 & 53.24 & 52.39 & 56.32 & 53.88 & 60.23 & \textbf{60.43} & \textbf{60.42} & 58.15 & 58.17 & 58.37 \\
RepoBench-P & 46.11 & \textbf{60.71} & 60.59 & 58.33 & 57.04 & \textbf{62.81} & 60.59 & 60.86 & 60.51 & \textbf{64.95} & 62.97 & 62.01 & 61.04 \\
\midrule
Average & 34.77 & 32.29 & 32.27 & 35.31 & \textbf{36.34} & 32.59 & 32.27 & 40.88 & \textbf{42.47} & 36.18 & 35.76 & 43.98 & \textbf{45.54} \\
\bottomrule
\end{tabular}
\label{tab:llama3-long}
}
\caption{
    \label{tab:llama3}
    The results comparison based on \basellama~\cite{llama3}. Our results are highlighted in teal and best results are indicated in bold.
}
\end{table*}

\subsection{Memory Block} 
\label{sec:block}
In light of the local semantic coherence in extended sequences, referring to previous studies~\cite{infllm}, we perform a memory lookup at the block level. We segment the context tokens \(\mathbf{C}\) into multiple memory blocks, each containing \(l_{b}\) tokens. We then select \(n_r\) tokens that have the highest representative scores to represent the block. For the \(i\)-th token, the representative score is calculated as 
\begin{equation}
    r_i = \frac{1}{l_L}\sum_{j=1}^{l_L}\mathbf{q}_{i+j}\cdot \mathbf{k}_i,
\end{equation}
where $l_L$ is the length of local token, \(\mathbf{q}_{i+j}\) is the query vector for the \((i+j)\)-th token and \(\mathbf{k}_i\) is the key vector for the \(i\)-th token. The score \(r_i\) intuitively measures the importance of the \(i\)-th token within its local window, demonstrating its influence on other tokens in the same window. 

\subsection{\lookup}
\label{sec:lookup}
When humans read and comprehend text, they first read the question and then search for the answer within the context with the question in mind. For instance, in \cref{fig:example}, when reading a novel with the question \emph{"What nationality is Ruth Anvory?"}, we can quickly locate the query-related memory context, which is \emph{"\dots that she was American"}. Building on this concept, we introduce \lookup, a simple but efficient lookup strategy. Our defined criterion for search is to locate tokens relevant to the query. We propose the relevance score between a memory block \(\mathbf{B}\) and query tokens \(\mathbf{Q}\) as follows:
\begin{equation}
    s(\mathbf{B}, \mathbf{Q}) = \sum_{i=1}^{l_Q}\sum_{j=1}^{r_k} \mathbf{Q_q}_{i}\cdot \mathbf{B_k}_{j}^r,
\end{equation}
where $l_Q$ is the length of query tokens. $\mathbf{Q}_{q_i}$ is the $i$-th query vector of $\mathbf{Q}$ and $\mathbf{B_k}_{j}^r$ is the $j$-th representative key vector of $\mathbf{B}$. The score \(s(\mathbf{B}, \mathbf{Q})\) is independent of the current token \(\mathbf{H}\), therefore, it only needs to be calculated once during inference.

On the other hand, the importance of different memory blocks is influenced by varying current tokens~\cite{infllm}. Therefore, the relevance score with the current token is also a criterion for selecting a memory block. The relevance score between a memory block $\mathbf{B}$ and current tokens $\mathbf{H}$ is defined as:
\begin{equation}
    s(\mathbf{B}, \mathbf{H}) = \sum_{i=1}^{l_H}\sum_{j=1}^{r_k} \mathbf{H_q}_{i}\cdot \mathbf{B_k}_{j}^r,
\end{equation}
where $l_H$ is the length of current tokens. $\mathbf{H}_{q_i}$ is the $i$-th query vector of $\mathbf{H}$ and $\mathbf{B_k}_{j}^r$ is the $j$-th representative key vector of $\mathbf{B}$. The final memory block score is thus composed of these two components:
\begin{equation}
    s(\mathbf{B}) = s(\mathbf{B}, \mathbf{H}) + \beta s(\mathbf{B}, \mathbf{Q}),
\end{equation}
where \(\beta\) represents the balancing factor. We opt to store the $n_b$ memory blocks with the highest scores in the current key-value cache. In terms of intuition, \(s(\mathbf{B}, \mathbf{Q})\) and \(s(\mathbf{B}, \mathbf{H})\) respectively express the search for the memory blocks related to the query tokens \(\mathbf{Q}\) and the current tokens \(\mathbf{H}\). Ablation experiments in \cref{sec:ablation} validate that \(\beta \geq 1\), indicating that the selection of queries is more crucial to the memory block. This aligns with our initial motivation. More methodology details are in \cref{sec:a-method}.

\begin{figure*}[t]
  \centering
  
  
  \includegraphics[width=\linewidth, trim=0 10 0 10, clip]{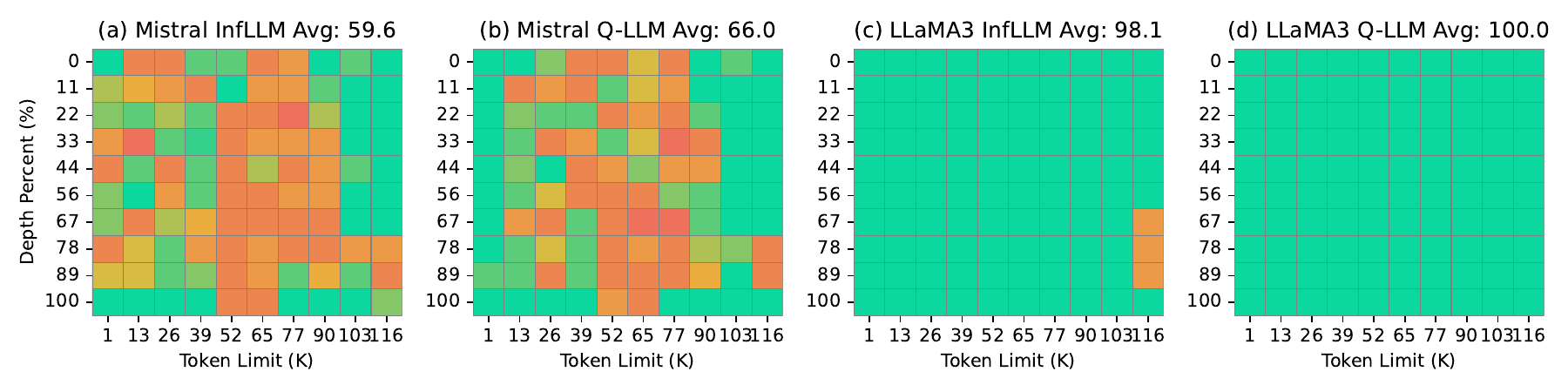}
\caption{The comparison of performance in the \needle task. The horizontal axis represents the document's length (the \emph{haystack}), whereas the vertical axis specifies the location of a brief sentence (the \emph{needle}) within the document, ranging from 1K to 128K tokens. A red cell indicates the language model's inability to recall the needle's information, while a green cell denotes successful recall by the model.}
\label{fig:needle}
\end{figure*}

\section{Experiments}
\label{sec:exp}

In this section, we conduct experiments utilizing \basemistral~\cite{mistral} and \basellama~\cite{llama3} as our base models. We compare our methods with \llamagrad (LLaMA-1048K)~\cite{llama3-1048k} and other competing sliding window approaches, containing LM-Infinite (Infinite)~\cite{infinite-llm}, StreamingLLM (Stream)~\cite{stream-llm} and InfLLM (Infllm)~\cite{infllm}. We test the methods on three cache lengths: 512, 1024 (1K), and 2048 (2K). More configuration details are in \cref{sec:a-conf}. Note that we add the queries before the context  to ensure the baselines also have query-aware capabilities, as detailed in \cref{sec:a-prompts}.

\begin{figure*}[t]
  \centering
\includegraphics[width=\linewidth, trim=10 10 10 10, clip]{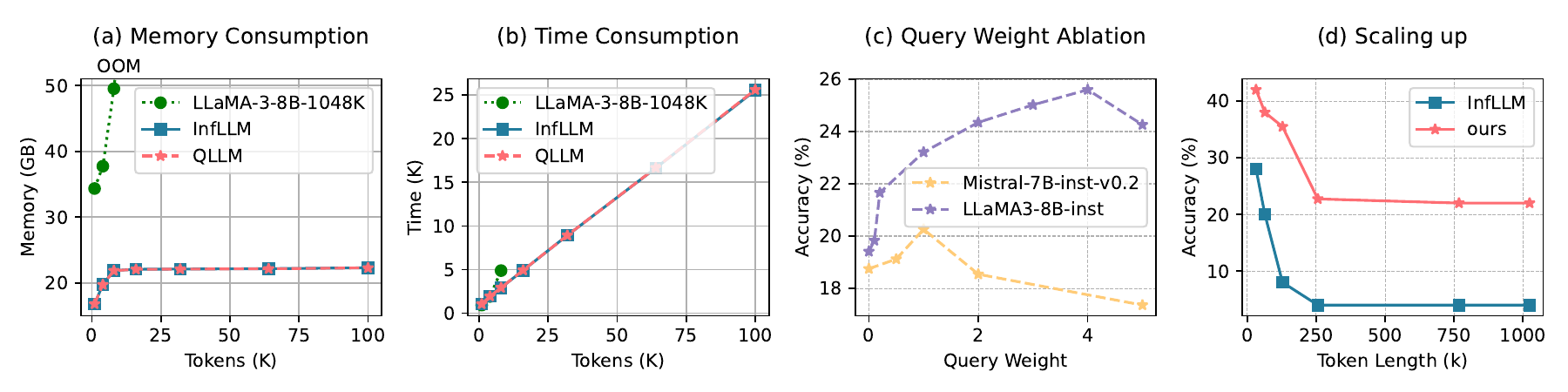}
\caption{(a) Memory and (b) Time consumption of different methods as tokens increases. (c) Ablation of query weight $\beta$. (d) The results of methods on sequences with extremely lengthy sequence lengths.}
\label{fig:exp}
\end{figure*}

\subsection{Long-Bench and $\infty$-Bench}
\label{sec:long_result}

In this section, we utilize representative tasks from two widely-recognized long document benchmarks, $\infty$-Bench~\cite{infinitebench} and Long-Bench~\cite{longbench}.  
The $95$\% quantile for sequence lengths in $\infty$-Bench and Long-Bench reaches $214$K and $31$K tokens. 
The outcomes based on \basemistral and \basellama are detailed in \cref{tab:mistral} and \cref{tab:llama3} respectively. The following observations can be made from the results: (1) Our approach shows considerable enhancement in performance when compared to base models (\llamagrad) and that utilizing the sliding window mechanism (StreamingLLM and LM-Infinite) across benchmarks and context window lengths. This suggests that the context memory in \name can effectively provide LLMs with appropriate contextual data, facilitating efficient comprehension and reasoning on long sequences. (2) Our method also exhibits a significant performance uplift when compared to models with other lookup mechanisms (InfLLM). This implies that previous methods still struggle to extract valid information from noisy contexts. Our proposed \lookup, however, can purposefully use the query to find relevant information in the long context. (3) Our technique is particularly beneficial in scenarios with longer input contexts and relatively smaller available context windows, as observed in comparisons across different benchmarks and context budgets. For instance, with a context window of 512 on the $\infty$-bench, \name improved by 7.17\% compared to the current SOTA on LLaMA3, and by 3.26\% on Mistral. This illustrates our model's superiority in infinite stream scenarios. (4) Our model's relative improvement is more noticeable on LLaMA3 when compared to other models. This is because superior models can more effectively utilize query information to precisely locate relevant information in the long context. 

\subsection{Needle in a Haystack} 
\label{sec:needle_result}
\textbf{\needle}~\cite{needleinhaystack} is a widely used benchmark to evaluate if models can effectively utilize extended context lengths. This test requires the model to accurately reproduce the details from a specific sentence (\emph{needle}) that is randomly positioned within a document that could be as long as 128K (\emph{haystack}). The results for methods based on \basemistral and \basellama are shown in \cref{fig:needle}. The context window size is 512. Our method accurately locates the \emph{needle} within the \emph{haystack} across 1K to 128K tokens. Specifically, \name improved upon InfLLM by 7.0\% on Mistral and achieved 100\% on LLaMA3. 

\vspace{2mm}\noindent\textbf{BABILong}~\cite{babilong} is a challenging benchmark for evaluating the performance of models in processing arbitrarily long documents with distributed facts. We conducted experiments on the BABILong based on \basellama. At a window length of 1024, our method improved by 6.1\% compared to InfLLM.

\begin{figure}[t]
  \centering
\includegraphics[width=0.9\linewidth, trim=5 5 5 5, clip]{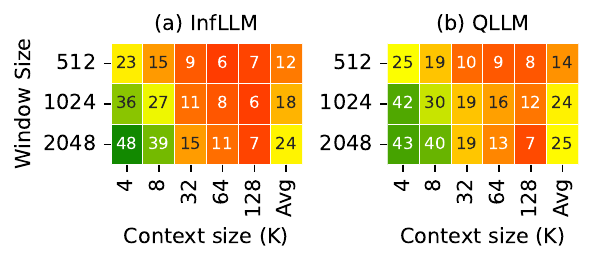}
\caption{Average accuracy over QA1-QA5 tasks from BABILong. The horizontal and vertical axes represent the context and window length. }
\label{fig:exp}
\end{figure}

\subsection{Time and Memory Consumption} 
\label{sec:speed}
In \cref{fig:exp}(a) and (b), we compare the time and memory consumption of different input tokens across methods. \name and InfLLM are comparable in terms of efficiency and memory usage. The time consumed by InfLLM and \name increases almost linearly with the number of input tokens, requiring only 25.6 seconds and 22.3GB of memory to process 100k tokens. In contrast, \llamagrad shows a rapid increase in time and memory consumption with the number of input tokens and cannot handle 16k tokens on a single A800 GPU (maximum memory of 80GB). The context length for InfLLM and \name is 2048.

\subsection{Ablation Experiments}
\label{sec:ablation}
To further substantiate the efficacy of \lookup, we carry out ablation studies in this section, with results displayed in~\cref{fig:exp}(c). The performance of \basemistral and \basellama exhibits a trend of initial increase followed by a decrease as the query weight $\beta$ escalates. We select the $\beta$ at the peak point as the experimental setup, choosing $\beta=1$ for Mistral and $\beta=4$ for LLaMA3. Further exploration of the primary elements within the context memory is in \cref{sec:a-explore}.


\subsection{Scaling up}
\label{sec:scale}
In this sub-section, we're evaluating \name's capacity to handle extremely lengthy sequences by extending the sequence length to $1024$K. The base model used is \basemistral and the task is \emph{Retrieve.KV} from $\infty$-Bench. The outcomes are displayed in~\cref{fig:exp}(d). According to the results, even when the context length is scaled to $1024$ thousand tokens, \name consistently performs at a level significantly above the current state-of-the-art. This confirms \name's ability to accurately recognize long-distance dependencies for effective long-sequence reasoning.

\section{Conclusion}
In this study, we focused on the significant challenges faced by LLMs in processing and reasoning over extensive contexts. We introduced \name, an approach inspired by human cognitive processes, which focuses on relevant memory data and effectively bypasses context input clutter. Our method does not require additional training and can be seamlessly integrated with any LLM. Through comprehensive evaluations using the LLaMA and Mistral models on the Longbench, $\infty$-Bench, and \needle benchmarks, \name demonstrated a marked improvement over the current SOTA. Moreover, our \llama can read 100K tokens within 30 seconds. The empirical results validate \name's ability to capture long-range dependencies and manage vast contexts efficiently, paving the way for enhanced performance in LLM-driven tasks that require long-sequence reasoning.

\section*{Limitations}
\label{sec:limit}
While \name demonstrates promising improvements over the current SOTA methods in various benchmarks, there are still some limitations. For instance, the system's performance relies on the limited window size, which could lead to potential information loss when dealing with highly complex contexts. Future research should address these limitations and explore the potential of \name in a broader range of tasks and contexts.

\section*{Broader Impact}
\label{sec:broader_impact}
The advancements made by \name in understanding and reasoning over broad contexts, a long-term research focus of Large Language Models (LLMs), could have profound implications across various fields. Given its ability to manage lengthy sequences,\name's potential to operate consistently over the content of conversations spanning recent days could make ChatBot assistants more effective and user-friendly. Tasks such as summarizing and answering questions based on books, reports, and documents, as well as generating code at the repository level, could also be improved with the ability to handle long context sequences. However, it is crucial to recognize that the benefits of \name also come with potential risks. The system's ability to process and understand extensive contexts could be misused for nefarious purposes, such as creating deepfakes or other forms of misinformation.

\bibliography{custom}

\appendix
\section{Methodology Details}
\label{sec:a-method}
In this section, we introduce our methodology details.
\subsection{Chunks}
Given the constraints of GPU memory, we do not encode the input sequence at once~\cite{infllm}; instead, we process it in chunks and generate output on a token-by-token basis. 
In each computational step, the inputs are composed of past key-value vectors \(\mathbf{P} = \{(\mathbf{k}_i, \mathbf{v}_i)\}_{i=1}^{l_P}\) and current token hidden vectors \(\mathbf{H}=\{\mathbf{h}_i\}_{i=1}^{l_H}\). 
When encoding, \(l_H\) is equivalent to the chunk size, while during decoding, \(l_H\) is equal to one.

\subsection{Positional Encoding}
Traditional LLM training typically utilizes a limited set of positional encodings, which can face difficulties with out-of-domain distribution when extended to process longer sequences~\cite{infllm}. Furthermore, in \name, the current key-value cache consists of several discontinuous text blocks. Assigning continuous positional encodings to these blocks could create mismatches and confuse the model. Consequently, drawing upon previous studies~\cite{t5,rerope2023,infllm}, we assign identical positional encodings to all tokens exceeding the local window size. More precisely, we set the distance between tokens in context memory blocks and the current tokens as \(l_L\).

\subsection{Cache Management} 
\label{sec:cache}
In order to process exceedingly lengthy sequence streams and encapsulate the semantic relevance with LLMs~\cite{infllm}, it's necessary to maintain all memory blocks and reference them at every computational stage. Given that most blocks are seldom used, we adopt an offloading strategy, which stores most memory blocks in CPU memory. Only the tokens and memory blocks essential for current operations are kept in GPU memory. Furthermore, due to the semantic continuity in long sequences where neighboring tokens often necessitate similar memory blocks, we assign a cache area in GPU memory, governed by a least recently used policy. This method enables efficient encoding of exceptionally long sequences using finite GPU memory. Moreover, for extremely long sequences, the representative tokens for each block can be offloaded to CPU memory, forming an effective k-nearest-neighbor index, thereby further diminishing computational complexity.

\section{Further Exploration}
\label{sec:a-explore}

\begin{figure*}[t]
  \centering
    \subfloat[Representative Tokens]{\includegraphics[width=0.33\linewidth, trim=0 0 0 0, clip]{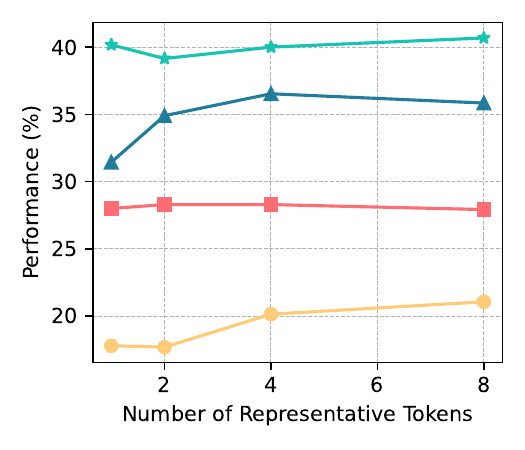}\label{fig:repr}}
    \subfloat[Block Size]{\includegraphics[width=0.33\linewidth, trim=0 0 0 0, clip]{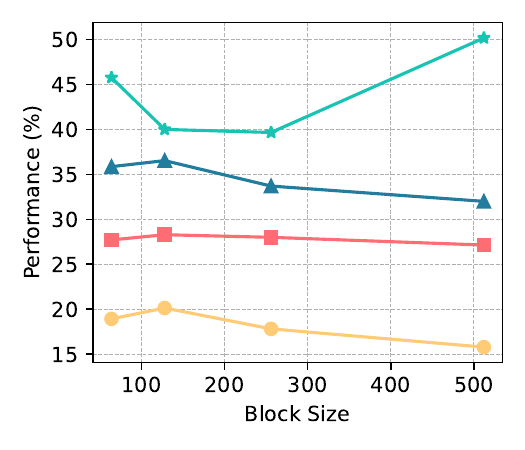}\label{fig:bs}}
    \subfloat[Number of Blocks]{\includegraphics[width=0.33\linewidth, trim=0 0 0 0, clip]{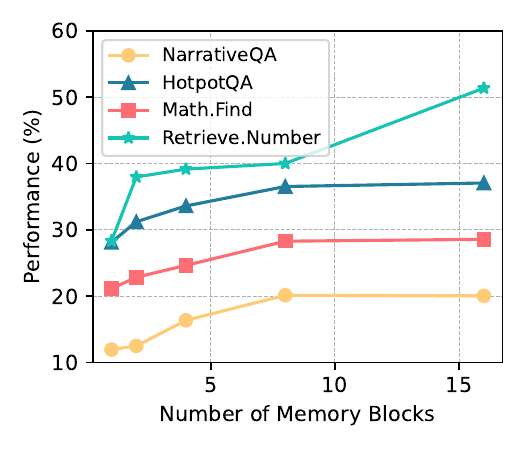}\label{fig:topk}}
    \caption{Further exploration, investigating the influence of the context memory with varying numbers of representative tokens, selected memory blocks, and memory block sizes, respectively.}
  \label{fig:hyper-parameter}
\end{figure*}

\name leverages context memory to retrieve pertinent data. We delve deeper into the influence of primary elements within the context memory. Results are presented in ~\cref{fig:hyper-parameter}. Conduct experiments on Mistral-7B-inst-v0.2 using the default parameters with a context window length of 1024.

\subsection{Number of Representative Tokens}
\name divides key-value vectors into memory blocks and picks a few representative tokens from each block to act as the segment's representation. The capacity of these tokens to symbolize the entire segment semantically directly impacts the model's efficacy. We run tests with the different quantity of representative tokens. The outcomes are displayed in~\cref{fig:repr}. We note an upward trend in model performance as the number of tokens increases, suggesting that a larger token count can better capture the semantic essence of memory segments. However, when the token count hits $8$, a slight performance dip is seen in HotpotQA. This drop can be traced back to the inclusion of semantically unrelated tokens in the segment representations. Future work could enhance model performance by developing more efficient and potent segment representations.

\subsection{Memory Block Size}
Each memory block should ideally represent a consistent semantic block. Oversized blocks can obstruct precise lookup, while undersized ones can escalate the computational cost of memory lookup. We test \name with different block sizes while maintaining a total context length of $1024$. The outcomes are displayed in ~\cref{fig:bs}. It's evident that the best block size changes according to the task due to the differing characteristics of input sequences. For instance, in NarrativeQA, a semantically complete paragraph forms a semantic block, while in Retrieve.Number, a single number does. Using heuristic rules to segment context can result in less-than-optimal performance. Thus, investigating how to dynamically segment context is a vital future research direction.

\subsection{Number of Memory blocks}
The chosen blocks are employed to supply relevant context to LLMs. We run tests with different number of blocks. 
From \cref{fig:topk}, we see that as the number of chosen blocks rises, there's a significant improvement in model performance. This is because more blocks mean a higher recall rate of relevant content. However, a larger block count also increases the time needed for memory scheduling and the computational time for attention. Hence, advancing lookup accuracy continues to be a key area for enhancing the efficiency of LLMs.

\section{Configuration Details}
\label{sec:a-conf}

\begin{figure*}[t]
  \centering
    \includegraphics[width=\linewidth, trim=0 0 60 0, clip]{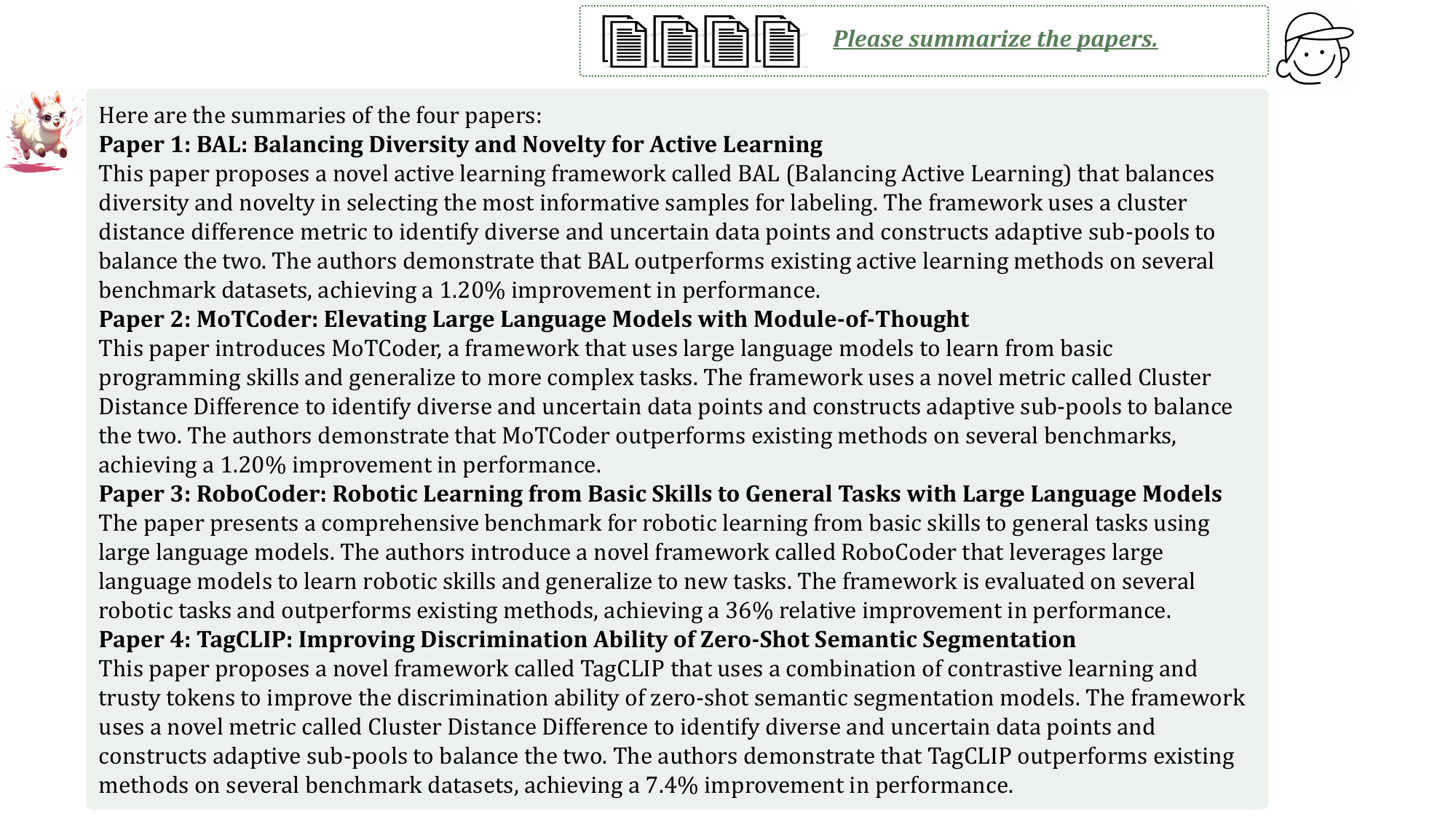}
    \caption{Examples of our \llama-8B summarizing multiple papers~\cite{bal,motcoder,tagclip,robocoder}. }
    \label{fig:a-exp-sum-papers}
\end{figure*}

\begin{table}[t]
  \centering
\small
\setlength{\tabcolsep}{1mm}
\begin{tabular}{cccc}
\toprule
Context Window & Local Tokens & Block Size & Block Num \\
\midrule
512 & 256 & 64 & 4 \\
1024 & 512 & 64 & 8 \\
2048 & 1024 & 128 & 8 \\
\bottomrule
\end{tabular}
\caption{
    \label{tab:conf}The parameters for different context window length, including number of local tokens, memory block size and number of memory blocks.
}
\end{table}

\subsection{Datasets}
\label{sec:a-dataset}
We utilize representative tasks from following widely-recognized long document benchmarks.

\vspace{2mm}\noindent(1) \textbf{$\infty$-Bench~\cite{infinitebench} and LongBench~\cite{longbench}}. Given that our base models are primarily pre-trained on English corpora, we employ English datasets for the evaluation. These benchmarks encompass a variety of tasks such as question answering, summarization, few-shot learning, context retrieval, mathematical computing, and code completion. The average document length in $\infty$-Bench is $145.1$K tokens, and in LongBench, it is $12.8$K tokens. The $95$\% quantile for sequence lengths in these benchmarks reaches $214$K and $31$K tokens respectively, which significantly exceeds the maximum length of the base models.

\vspace{2mm}\noindent (2) \textbf{Needle-in-a-Haystack}~\cite{needleinhaystack}, a widely used benchmark to evaluate if models can effectively use extended context lengths. This test requires the model to accurately reproduce the details from a specific sentence (referred to as the \emph{needle}) that is randomly positioned within a document that could be as long as 128K (referred to as the \emph{haystack}). We adopted the following setting: the needle is \textit{The best thing to do in San Francisco is eat a sandwich and sit in Dolores Park on a sunny day.}, and the haystack is \textit{PaulGrahamEssays}. The retrieval question is \textit{What is the best thing to do in San Francisco?}. 

\vspace{2mm}\noindent(3) \textbf{Scaling up}. To evaluate \name's capacity to handle extremely lengthy sequences by extending the sequence length to $1024$K. We use the \emph{Retrieve.KV} task from the $\infty$-Bench to test its ability to discern context in extensive sequences. This task requires LLMs to identify a value from a key and a dictionary, essentially locating pertinent information within long sequences. Inputs with $\{32, 64, 128, 256, 768, 1024\}$ thousand tokens are automatically generated.
For each length, 50 instances are created for assessment. 

\vspace{2mm}\noindent(4) \textbf{BABILong} is a challenging benchmark for evaluating the performance of models in processing arbitrarily long documents with distributed facts. BABI tasks are generated by simulating a set of characters and objects engaged in various movements and interactions with each other in multiple locations. Each interaction is represented by a fact, and the task is to answer a question using the facts from the current simulation. The bAbI tasks vary based on the number of facts, question complexity and the aspects of reasoning.

\subsection{Baseline Methods}
Our goal is to enable LLMs trained with limited sequence lengths to comprehend extremely long sequences without additional training. For this purpose, we use Mistral-7B-Instruct-v0.2~\cite{mistral} and LLaMA3-8B-inst~\cite{llama3} as our base models. Mistral-7B-Instruct-v0.2 is initially pre-trained with a maximum sequence length of $8$K tokens and subsequently fine-tuned with a maximum sequence length of $32$K tokens. LLaMA3-8B-inst is fine-tuned from LLaMA3-8B~\cite{llama3} with an $8$K token maximum length. We compare our methods with competing sliding window approaches, containing LM-Infinite (Infinite)~\cite{infinite-llm}, StreamingLLM (Stream)~\cite{stream-llm} and InfLLM (infllm)~\cite{infllm}. The same prompts are utilized for all methods and detailed in \cref{sec:a-prompts}. 

\subsection{Implementation Details}
\label{sec:imple}
Following the settings of previous works~\cite{infllm}, the number of representative tokens, $r_k$, is set to 4. The number of initial tokens is set to 128 for LM-Infinite, StreamingLLM, InfLLM, and \name, to cover system prompts and task descriptions. We test the methods on three cache lengths: 512, 1024 (1K), and 2048 (2K). The local tokens, block size, and number of blocks for InfLLM and \name are detailed in~\cref{tab:conf}.

\section{More Examples}
\label{sec:a-examples}
\begin{figure*}[t]
  \centering
    \includegraphics[width=0.99\linewidth, trim=0 120 60 0, clip]{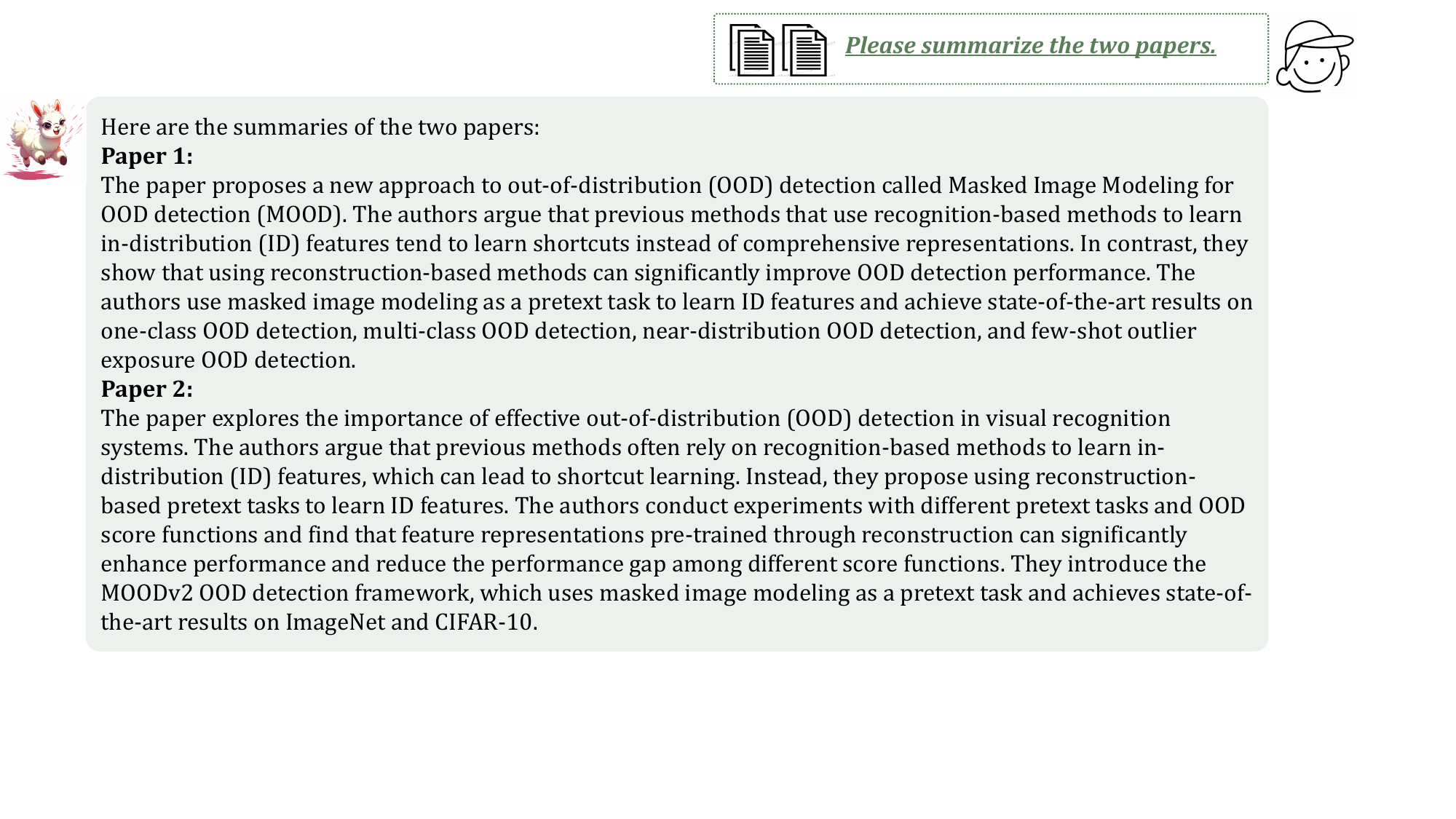}
    \\
    \includegraphics[width=0.99\linewidth, trim=0 140 60 0, clip]{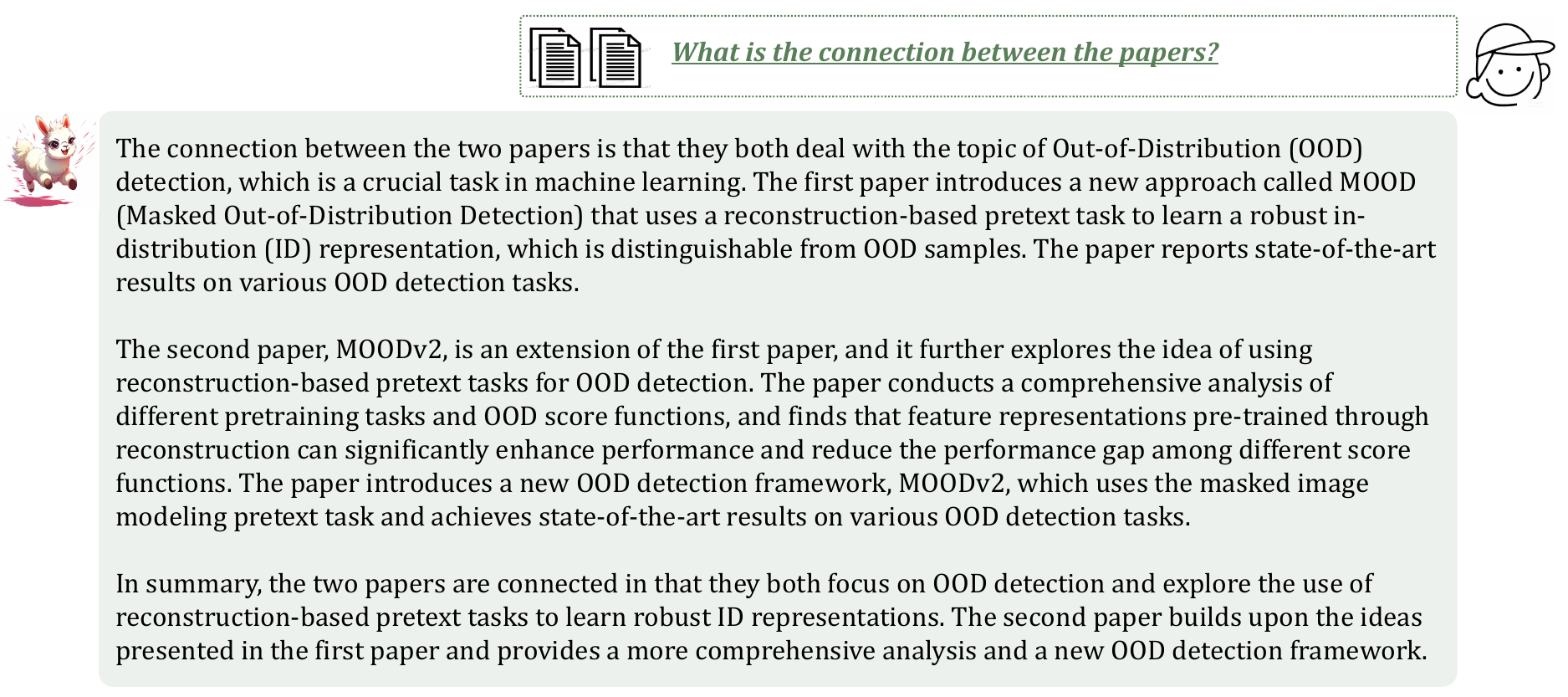}
    \\
    \includegraphics[width=0.99\linewidth, trim=0 0 60 0, clip]{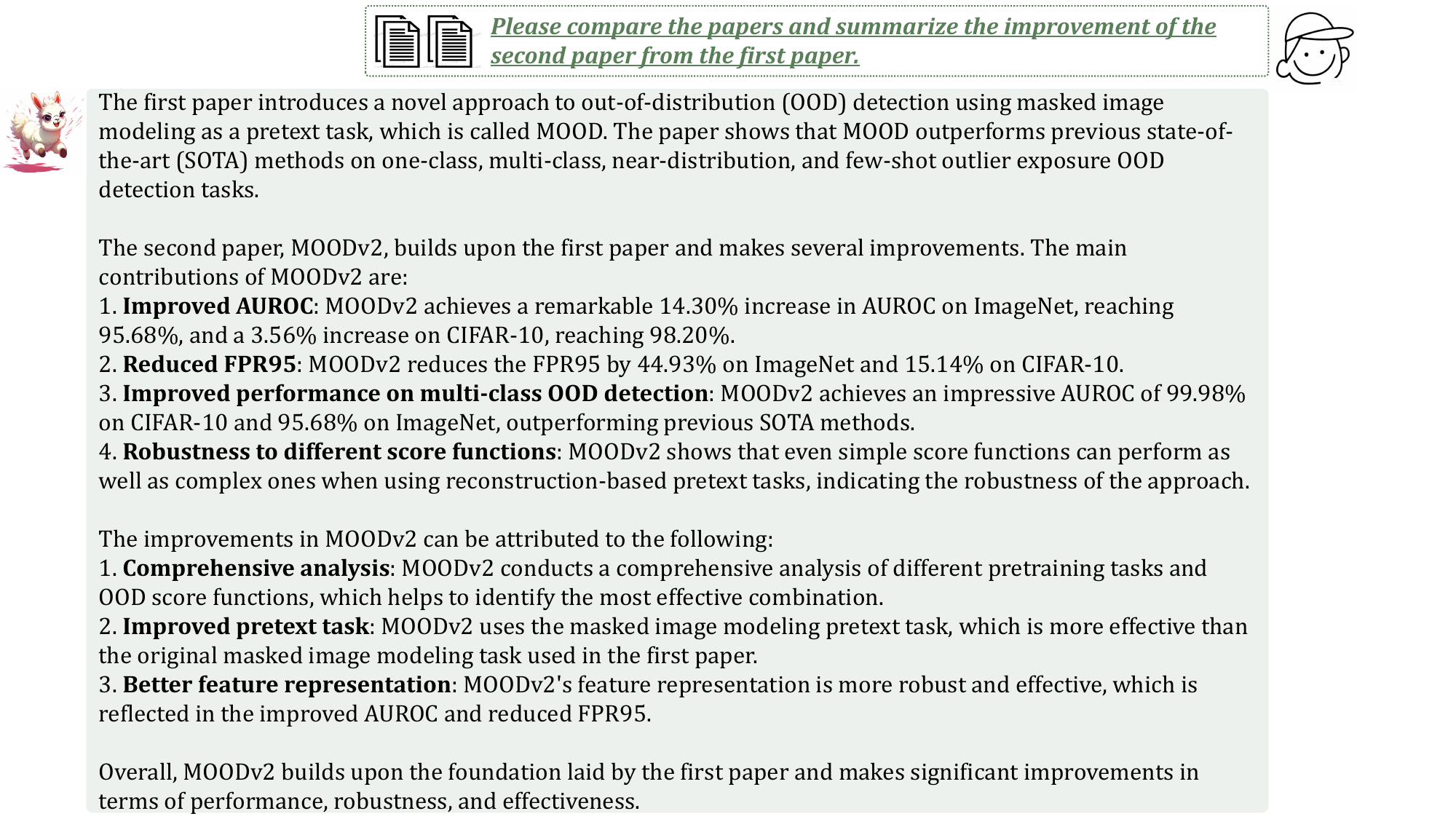}
    \caption{Examples of our \llama-8B summarizing paper series and discovering their connection and comparison~\cite{mood, moodv2}. }
    \label{fig:a-exp-mood}
\end{figure*}

\begin{figure*}[t]
  \centering
    \includegraphics[width=\linewidth, trim=0 100 60 0, clip]{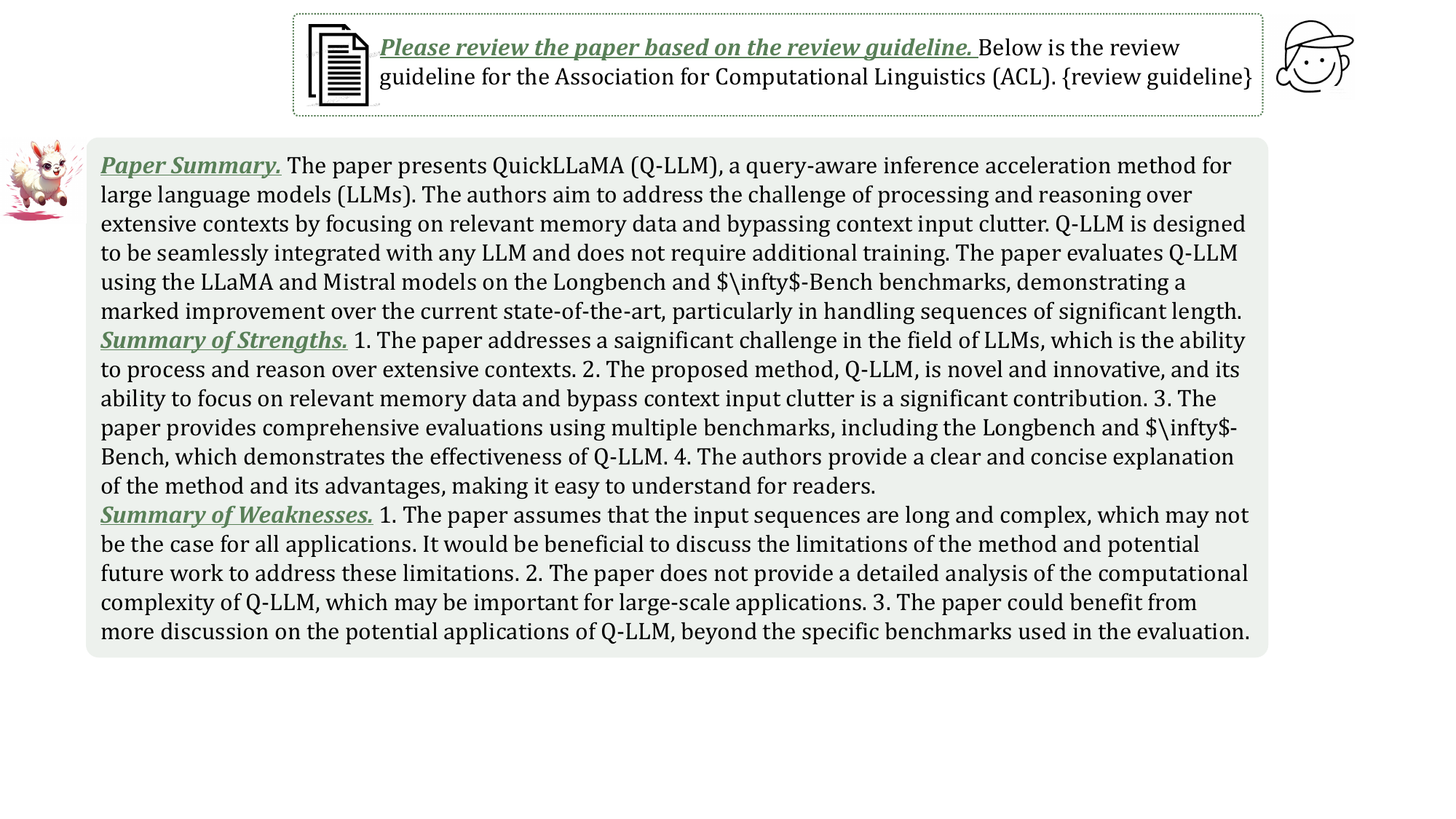}
    \includegraphics[width=\linewidth, trim=0 200 60 0, clip]{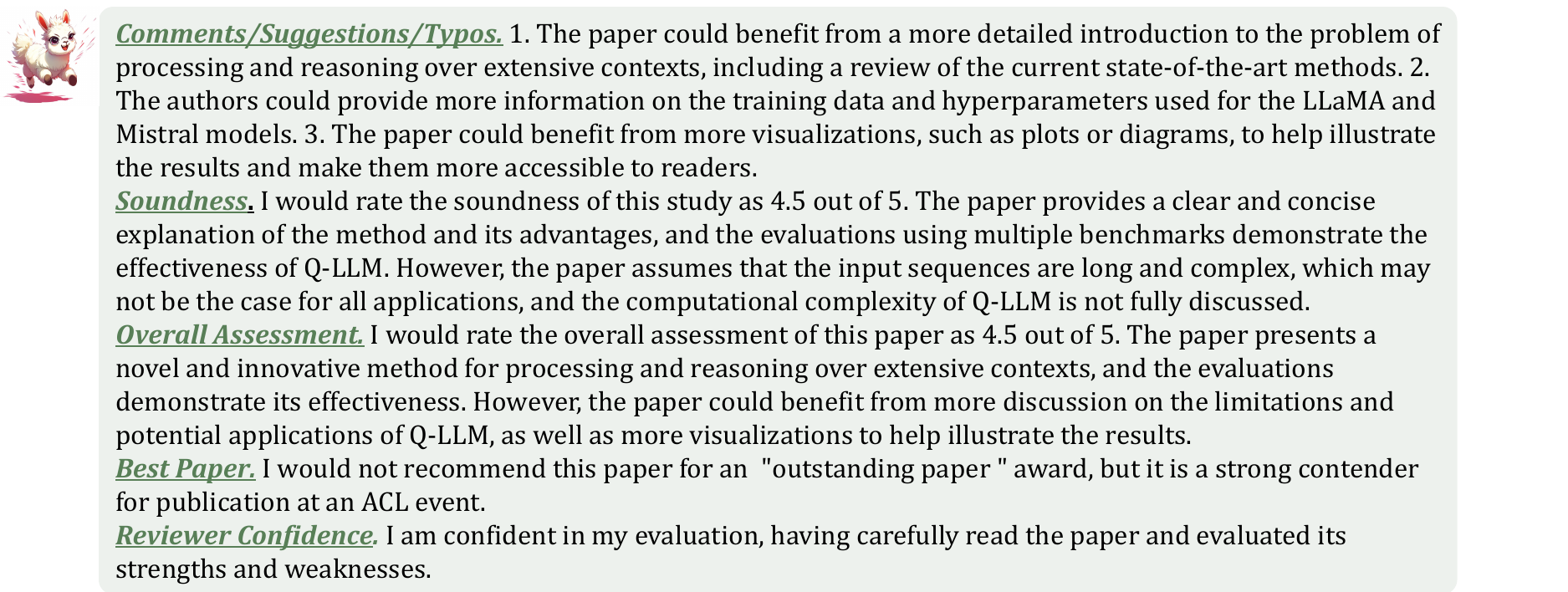}
    \caption{Examples of our \llama-8B reviewing our paper according to ACL review guideline. }
    \label{fig:a-exp-paper-review}
\end{figure*}

\begin{figure*}[t]
  \centering
    \includegraphics[width=\linewidth, trim=0 170 60 0, clip]{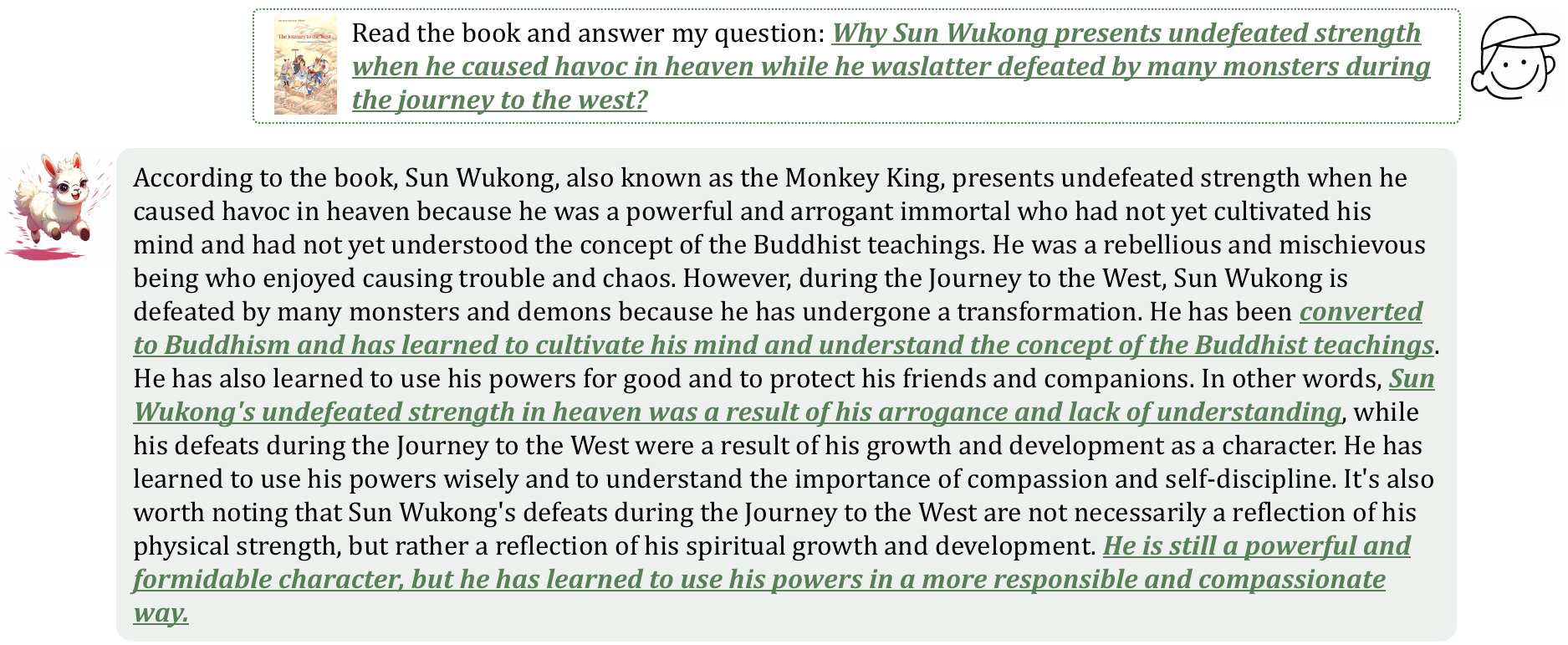}
    \includegraphics[width=\linewidth, trim=0 260 60 0, clip]{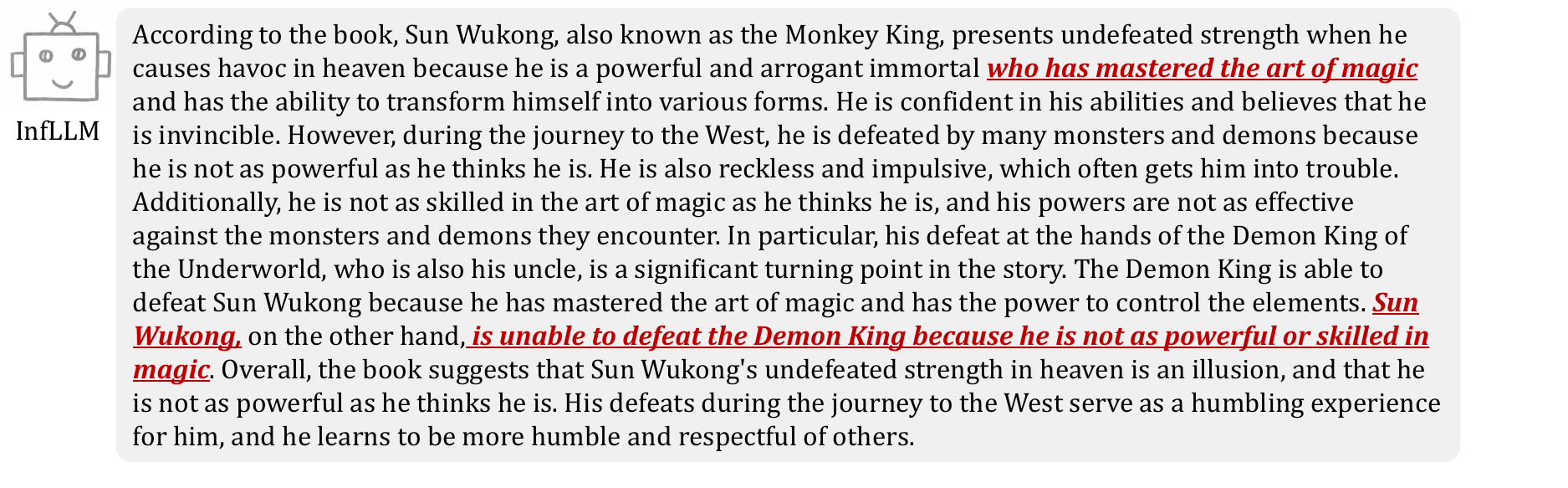}
    \includegraphics[width=\linewidth, trim=0 310 60 0, clip]{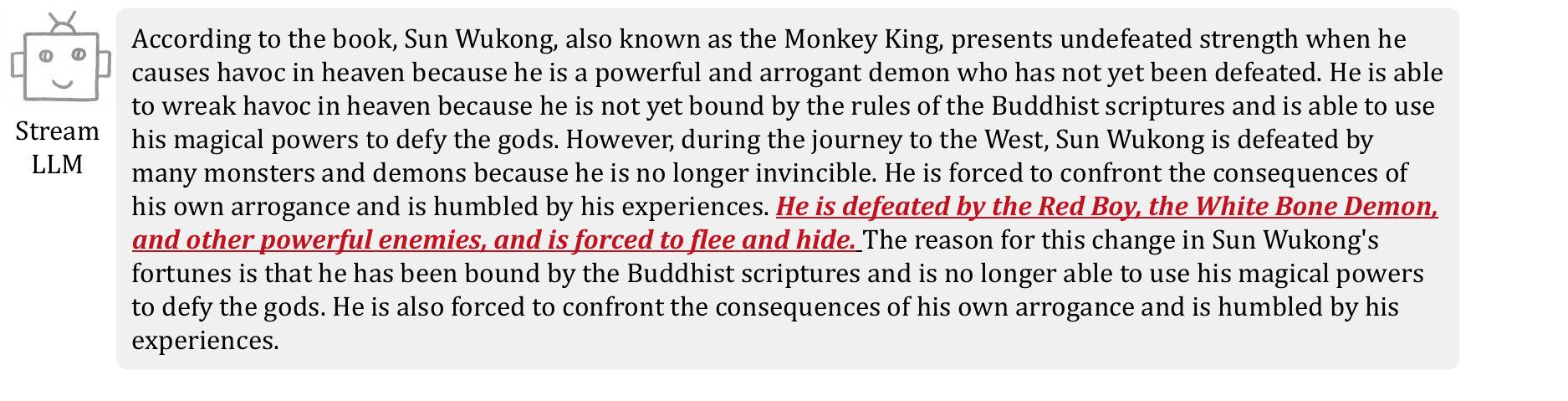}
    \includegraphics[width=\linewidth, trim=0 310 60 0, clip]{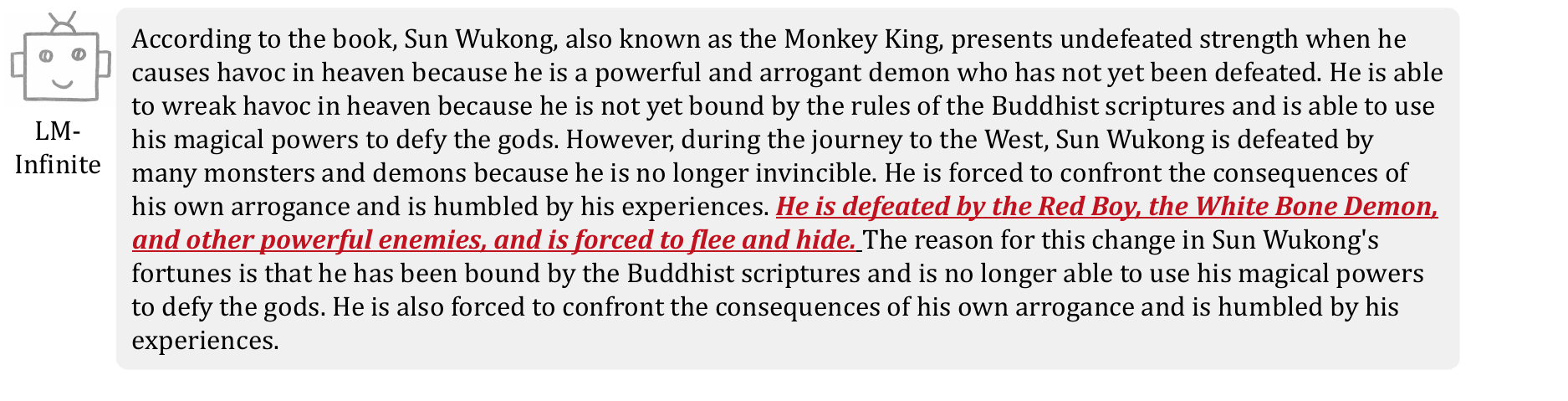}
    \caption{Examples of our \llama-8B reading long novel \emph{Journey to the West} comparing with current SOTAs}
    \label{fig:a-qualitive-com}
\end{figure*}

In this section, we provide examples of our \llama-8B.
\begin{enumerate}
    \item \Cref{fig:a-exp-sum-papers} summarizes multiple papers.
    \item \Cref{fig:a-exp-mood} summarizes paper series and discovering their connection and comparison.
    \item \Cref{fig:a-exp-paper-review} reviews our paper according to ACL review guideline. 
    \item \Cref{fig:a-qualitive-com} reads long novel \emph{Journey to the West} comparing with current SOTAs.
\end{enumerate}

\section{Prompts}
\label{sec:a-prompts}
We utilize the same prompts for all experiments in the main paper.
\begin{enumerate}
    \item The prompts for Long-Bench are in \cref{tab:prompt_long_1,tab:prompt_long_2}. 
    \item The prompts for $\infty$-Bench are in \cref{tab:prompt_infinite}.
    \item The prompts for \needle Bench are in \cref{tab:prompt_needle}.
\end{enumerate}

\begin{table*}[t]
  \centering
\small
\setlength{\tabcolsep}{1mm}
\begin{tabularx}{\textwidth}{l|X}
\toprule
Dataset & Prompt \\
\midrule
NarrativeQA & You are given a story, which can be either a novel or a movie script, and a question. Answer the question asconcisely as you can, using a single phrase if possible. Do not provide any explanation.\newline Question: \colorbox{yellow!30}{\{input\}}\newline Story: \colorbox{gray!25}{\{context\}}\newline Now, answer the question based on the story asconcisely as you can, using a single phrase if possible. Do not provide any explanation.\newline Question: \{input\}\newline Answer: \\
\midrule
Qasper & You are given a scientific article and a question. Answer the question as concisely as you can, using a single phrase or sentence if possible. If the question cannot be answered based on the information in the article, write "unanswerable". If the question is a yes/no question, answer "yes", "no", or "unanswerable". Do not provide any explanation.\newline Question: \colorbox{yellow!30}{\{input\}}\newline Article: \colorbox{gray!25}{\{context\}}\newline  Answer the question based on the above article as concisely as you can, using a single phrase or sentence if possible. If the question cannot be answered based on the information in the article, write "unanswerable". If the question is a yes/no question, answer "yes", "no", or "unanswerable". Do not provide any explanation.\newline Question: \{input\}\newline Answer: \\
\midrule
MultiFieldQA & Read the following text and answer briefly.\newline Question: \colorbox{yellow!30}{\{input\}}\newline \colorbox{gray!25}{\{context\}}\newline Now, answer the following question based on the above text, only give me the answer and do not output any other words.\newline Question: \{input\}\newline Answer: \\
\midrule
HotpotQA & Answer the question based on the given passages. Only give me the answer and do not output any other words.\newline The following are given passages.\newline Question: \colorbox{yellow!30}{\{input\}}\newline \colorbox{gray!25}{\{context\}}\newline Answer the question based on the given passages. Only give me the answer and do not output any other words.\newline Question: \{input\}\newline Answer: \\
\midrule
2WikiMQA & Answer the question based on the given passages. Only give me the answer and do not output any other words.\newline The following are given passages.\newline Question: \colorbox{yellow!30}{\{input\}}\newline \colorbox{gray!25}{\{context\}}\newline Answer the question based on the given passages. Only give me the answer and do not output any other words.\newline Question: \{input\}\newline Answer: \\
\midrule
Musique & Answer the question based on the given passages. Only give me the answer and do not output any other words.\newline The following are given passages.\newline Question: \colorbox{yellow!30}{\{input\}}\newline \colorbox{gray!25}{\{context\}}\newline Answer the question based on the given passages. Only give me the answer and do not output any other words.\newline Question: \{input\}\newline Answer: \\
\midrule
GovReport & You are given a report by a government agency. \colorbox{yellow!30}{Write a one-page summary of the report}.\newline Report:\newline \colorbox{gray!25}{\{context\}}\newline Now, write a one-page summary of the report.\newline Summary: \\
\bottomrule
\end{tabularx}

\caption{
    \label{tab:prompt_long_1}
    The prompt for each dataset in Long-Bench~\cite{longbench}. Yellow highlights indicate the query, while a gray background represents the long context.
}
\end{table*}

\begin{table*}[t]
\centering
\small
\setlength{\tabcolsep}{1mm}
\begin{tabularx}{\textwidth}{l|X}
\toprule
Dataset & Prompt \\

\midrule
QMSum & You are given a meeting transcript and a query containing a question or instruction. Answer the query in one or more sentences.\newline Query: \colorbox{yellow!30}{\{input\}}\newline Transcript:\newline \colorbox{gray!25}{\{context\}}\newline Now, answer the query based on the above meeting transcript in one or more sentences.\newline Query: \{input\}\newline Answer: \\
\midrule
MultiNews & You are given several news passages. \colorbox{yellow!30}{Write a one-page summary of all news}. \newline News:\newline \colorbox{gray!25}{\{context\}}\newline Now, write a one-page summary of all the news.\newline Summary: \\
\midrule
TREC & Please determine the type of the question below.\newline  \colorbox{yellow!30}{\{input\}} \newline Here are some examples of questions.\newline \colorbox{gray!25}{\{context\}}\newline Now please determine the type of the question below.\newline \{input\} \\
\midrule
TriviaQA & Answer the question based on the given passage. Only give me the answer and do not output any other words.\newline  \colorbox{yellow!30}{\{input\}} \newline The following are some examples.\newline \colorbox{gray!25}{\{context\}}\newline Now answer the question based on the given passage. Only give me the answer and do not output any other words. \newline \{input\} \\
\midrule
SAMSum & Summarize the dialogue into a few short sentences.\newline  \colorbox{yellow!30}{\{input\}} \newline The following are some examples.\newline \colorbox{gray!25}{\{context\}}\newline \{input\} \\
\midrule
PassageRetrieval & Here are 30 paragraphs from Wikipedia, along with an abstract. Please determine which paragraph the abstract is from. \newline Abstract:\newline  \colorbox{yellow!30}{\{input\}}\newline Paragraphs:\newline \colorbox{gray!25}{\{context\}}\newline Please enter the number of the paragraph that the abstract is from. The answer format must be like "Paragraph 1", "Paragraph 2", etc.\newline The answer is:  \\
\midrule
LCC & \colorbox{yellow!30}{Please complete the code given below}. \newline \colorbox{gray!25}{\{context\}}Next line of code:  \\
\midrule
RepoBench-P & \colorbox{yellow!30}{Please complete the code given below}. \newline \colorbox{gray!25}{\{context\}} \{input\}\newline Next line of code:  \\
\bottomrule
\end{tabularx}

\caption{
    \label{tab:prompt_long_2}
    The prompt for each dataset in Long-Bench~\cite{longbench} (continued). Yellow highlights indicate the query, while a gray background represents the long context.
}
\end{table*}

\begin{table*}[t]
\centering
\small
\setlength{\tabcolsep}{1mm}
\begin{tabularx}{\textwidth}{c|X}
\toprule
Dataset & Prompt \\
\midrule 
En.MC & Read the book and answer the question.Only one of the following options is correct, tell me the answer using one single letter (A, B, C, or D). Don't say anything else. \newline Question: \colorbox{yellow!30}{\{question\}}\newline A. \{OPTIONA\}\newline B. \{OPTIONB\}\newline C. \{OPTIONC\}\newline D. \{OPTIOND\}\newline \colorbox{gray!25}{\{context\}}\newline Question: \{question\}\newline Only one of the following options is correct, tell me the answer using one single letter (A, B, C, or D). Don't say anything else.\newline A. \{OPTIONA\}\newline B. \{OPTIONB\}\newline C. \{OPTIONC\}\newline D. \{OPTIOND\} \\
\midrule
Retrieve.PassKey & There is an important info hidden inside a lot of irrelevant text. Find and memorize it: \colorbox{yellow!30}{\{input\}}\newline \colorbox{gray!25}{\{context\}}\newline \{input\} \\
\midrule
Retrieve.Number & There is an important info hidden inside a lot of irrelevant text. Find and memorize it: \colorbox{yellow!30}{\{input\}}\newline \colorbox{gray!25}{\{context\}}\newline \{input\} \\
\midrule
Code.Debug & There is ONLY ONE function in the large project that is deliberately made to include an obvious error. Please find the function that contains the most obvious errors. I will give you four options to narrow your scope. You can inspect the options and think. Eventually, tell me the answer using one single letter (A, B, C, or D). \newline \colorbox{yellow!30}{Which funtion has deliberate error?}\newline A. \{OPTIONA\}\newline B. \{OPTIONB\}\newline C. \{OPTIONC\}\newline D. \{OPTIOND\}\newline \colorbox{gray!25}{\{context\}}\newline  Which funtion has deliberate error?\newline A. \{OPTIONA\}\newline B. \{OPTIONB\}\newline C. \{OPTIONC\}\newline D. \{OPTIOND\}\newline Give me your answer for the function that has the deliberate and obvious error in A, B, C, or D. Your answer MUST be chosen from one of the four options without any explanation. If you cannot determine answers accurately, you also MUST provide the answer you think is most likely. Absolutely do not say you do not know or you need more information. \\
\midrule
Math.Find & \colorbox{yellow!30}{\{prefix\}}\newline \colorbox{gray!25}{\{context\}}\newline \{input\} \\
\midrule
Retrieve.KV & Extract the value corresponding to the specified key \colorbox{yellow!30}{\{key\}} in the JSON object below.\newline \colorbox{gray!25}{\{context\}}\newline \{input\} \\
\bottomrule
\end{tabularx}

\caption{
    \label{tab:prompt_infinite}
    The prompt for each dataset in $\infty$-Bench~\cite{infinitebench} (continued). Yellow highlights indicate the query, while a gray background represents the long context.
}
\end{table*}

\begin{table*}[t]
\centering
\small
\setlength{\tabcolsep}{1mm}
\begin{tabularx}{\textwidth}{l|X}
\toprule
Dataset & Prompt \\
\midrule 
Needle in a Haystack & Based on the content of the book, Question: \colorbox{yellow!30}{What is the best thing to do in San Francisco?} <book> \colorbox{gray!25}{<context>} </book>.  \\
\bottomrule
\end{tabularx}

\caption{
    \label{tab:prompt_needle}
    The prompt for each dataset in Needle-in-a-Haystack Benchmark~\cite{needleinhaystack}. Yellow highlights indicate the query, while a gray background represents the long context.
}
\end{table*}

\end{document}